\documentclass[runningheads]{llncs}
\usepackage{graphicx}
\usepackage{amsmath,amssymb} % define this before the line numbering.
\usepackage{tabularx}
\usepackage{multirow}
\usepackage{subfigure}
\usepackage{dsfont}
\usepackage{colortbl}
\usepackage[table]{xcolor}
\usepackage{color}\usepackage[width=122mm,left=12mm,paperwidth=146mm,height=193mm,top=12mm,paperheight=217mm]{geometry}

\definecolor{dbcolor}{rgb}{1,0,0}

\newcommand{\figref}[1]{Fig. \ref{#1}}
\newcommand{\tabref}[1]{Table \ref{#1}}
\newcommand{\equref}[1]{(\ref{#1})}
\newcommand{\secref}[1]{Sec. \ref{#1}}

\makeatletter
\def\hlinewd#1{%
	\noalign{\ifnum0=`}\fi\hrule \@height #1 \futurelet
	\reserved@a\@xhline}

\begin{document}
\pagestyle{headings}
\mainmatter
\title{Deep Self-Convolutional Activations Descriptor for Dense Cross-Modal Correspondence} % Replace with your title

\titlerunning{Deep Self-Convolutional Activations Descriptor}

\authorrunning{S. Kim et al.}

\author{Seungryong Kim$^{1}$\thanks{This work is done while Seungryong Kim was an intern at Microsoft Research.}, Dongbo Min$^2$, Stephen Lin$^3$, and Kwanghoon Sohn$^1$}
\institute{$^1$Yonsei University, $^2$Chungnam National Univerisity, $^3$Microsoft Research}

\maketitle

\begin{abstract}
	We present a novel descriptor, called deep self-convolutional
	activations (DeSCA), designed for establishing dense correspondences
	between images taken under different imaging modalities, such as
	different spectral ranges or lighting conditions. Motivated by
	descriptors based on local self-similarity (LSS), we formulate a
	novel descriptor by leveraging LSS in a deep architecture,
	leading to better discriminative power and greater robustness to
	non-rigid image deformations than state-of-the-art cross-modality
	descriptors. The DeSCA first computes self-convolutions over a local support
	window for randomly sampled patches, and then builds self-convolution
	activations by performing an average pooling through a hierarchical
	formulation within a deep convolutional architecture.
	Finally, the feature responses on the self-convolution activations
	are encoded through a spatial pyramid pooling in a circular configuration.
	In contrast to existing convolutional neural networks (CNNs) based
	descriptors, the DeSCA is training-free (i.e., randomly sampled
	patches are utilized as the convolution kernels), is robust to
	cross-modal imaging, and can be densely computed in an efficient
	manner that significantly reduces computational redundancy. The
	state-of-the-art performance of DeSCA on challenging cases of
	cross-modal image pairs is demonstrated through extensive
	experiments.
\end{abstract}

\section{Introduction}\label{sec:1}
In many computer vision and computational photography applications,
images captured under different imaging modalities are used to supplement the data
provided in color images.
Typical examples of other imaging modalities include near-infrared \cite{Brown11,yan13,Hwang15}
and dark flash \cite{Krishnan09} photography.
More broadly, photos taken under different imaging conditions,
such as different exposure settings \cite{Sen12}, blur levels \cite{HaCohen13,Lee13},
and illumination \cite{Petschnigg04}, can also be considered as cross-modal \cite{Shen14,Kim15}.

Establishing dense correspondences between cross-modal image pairs
is essential for combining their disparate information.
Although powerful global optimizers may help to improve the accuracy of
correspondence estimation to some extent \cite{Liu11,Kim13},
they face inherent limitations without help of suitable matching descriptors \cite{Pinggera12}.
The most popular local descriptor is scale invariant feature transform (SIFT) \cite{Lowe04},
which provides relatively good matching performance when there are small photometric variations.
However, conventional descriptors such as SIFT often fail to capture reliable matching evidences
in cross-modal image pairs due to their different visual properties \cite{Shen14,Kim15}.

Recently, convolutional neural networks (CNNs) based features \cite{Simonyan14,Gong14,Fischer14,Donahue14,Simo-Serra15b}
have emerged as a robust alternative with high discriminative power.
However, CNN-based descriptors cannot satisfactorily deal with severe cross-modality appearance differences,
since they use shared convolutional kernels across images 
which lead to inconsistent responses similar to
conventional descriptors \cite{Simo-Serra15b,Dong15}. Furthermore,
they do not scale well for dense correspondence estimation due to
their high computational complexity. Though recent works \cite{Long15}
propose an efficient method that extracts dense outputs through the deep CNNs, they do not extract dense CNN features for all pixels individually. More
seriously, their methods were usually designed to perform a specific
task only, \emph{e.g.}, semantic segmentation, not to provide a general
purpose descriptor like ours.
\begin{figure}[!t]
	\centering
	\renewcommand{\thesubfigure}{}
	\subfigure[]
	{\includegraphics[width=0.65\linewidth]{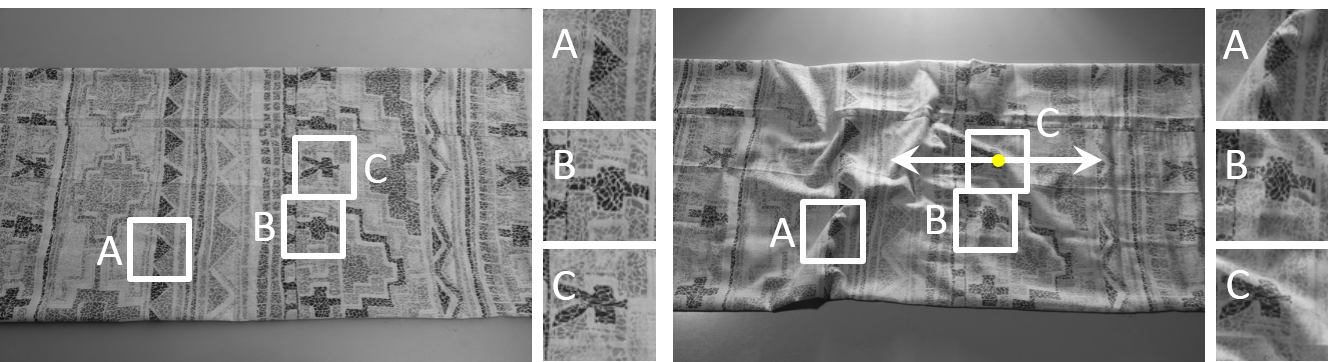}}\\
	\vspace{-18pt}
	\subfigure[(a) cost function in A]
	{\includegraphics[width=0.26\linewidth]{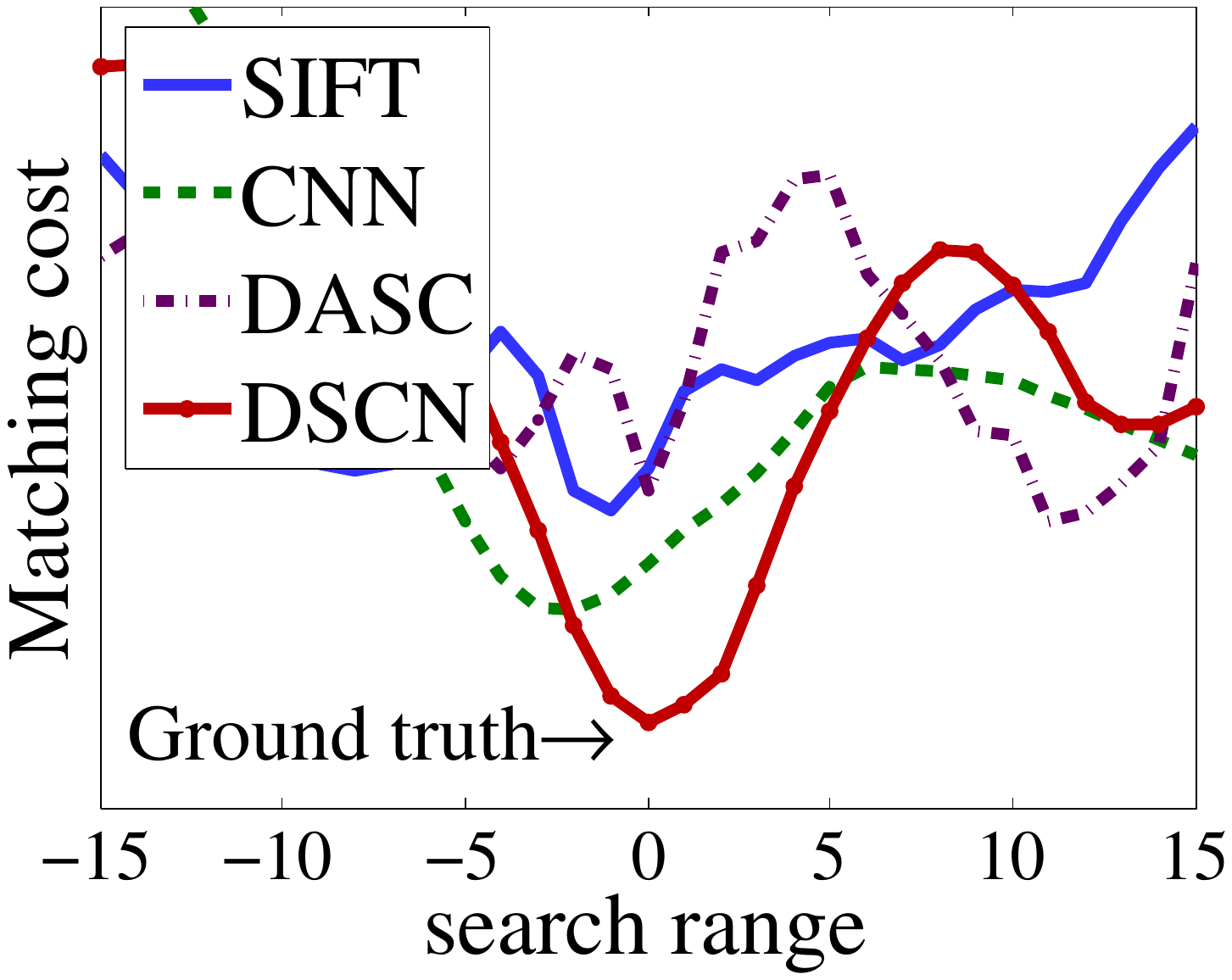}}
	\subfigure[(b) cost function in B]
	{\includegraphics[width=0.26\linewidth]{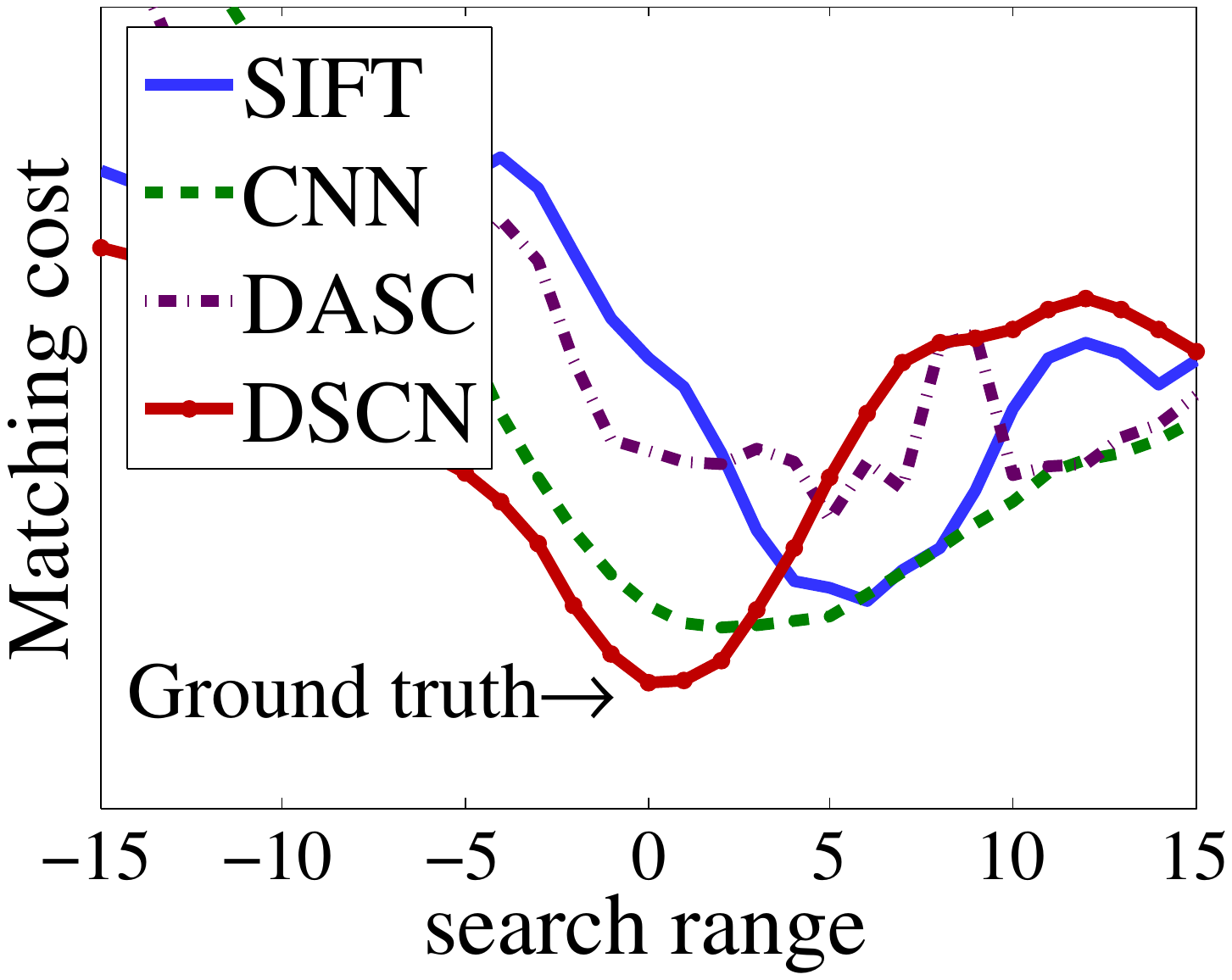}}
	\subfigure[(c) cost function in C]
	{\includegraphics[width=0.26\linewidth]{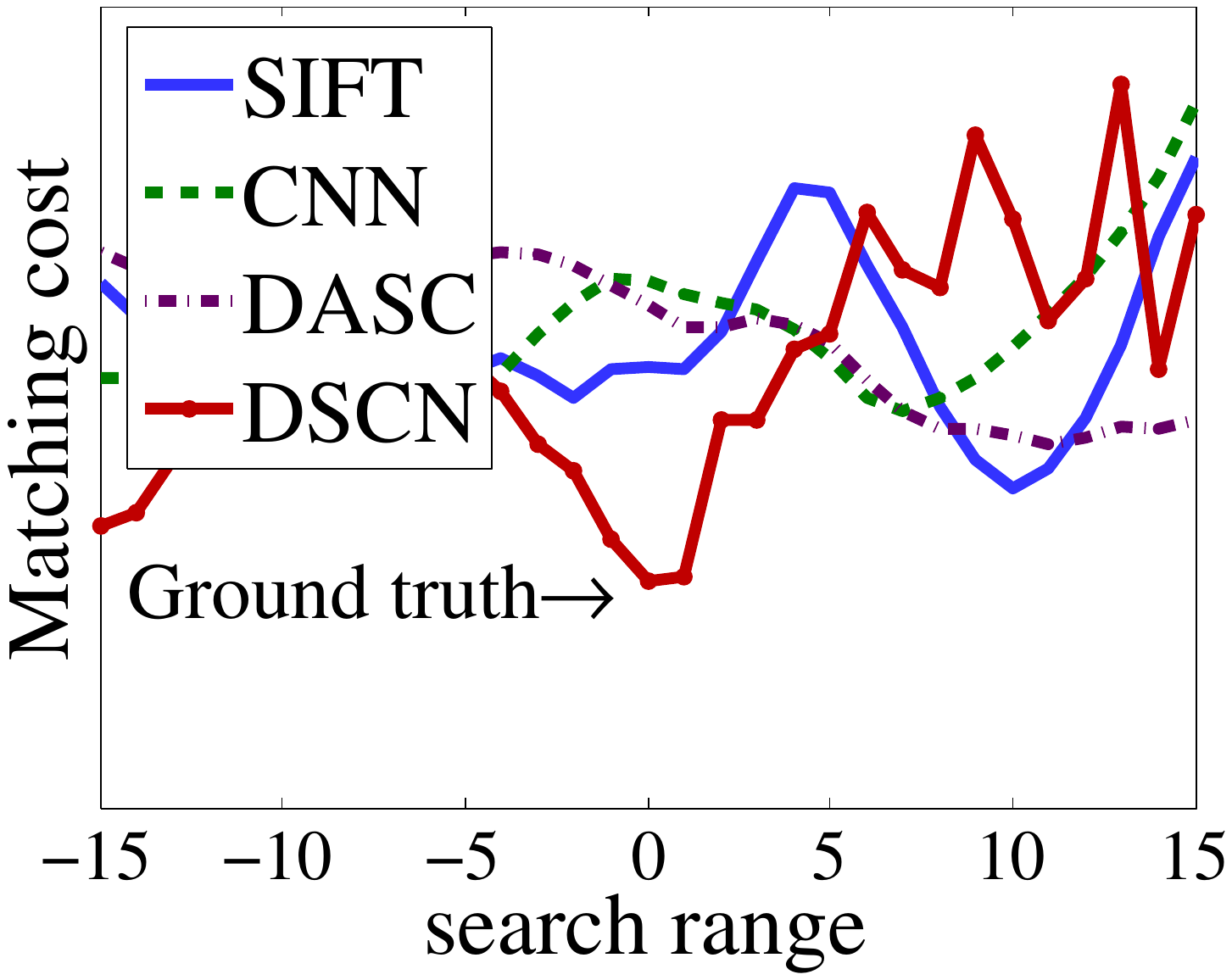}}\\
	\vspace{-12pt}
	\caption{Examples of matching cost profiles, computed with
		different descriptors along the scan lines of A, B, and C for image
		pairs under severe non-rigid deformations and illumination
		changes. Unlike other descriptors, DeSCA yields reliable
		global minimum.}\label{img:1}\vspace{-10pt}
\end{figure}

To address the problem of cross-modal appearance changes, feature
descriptors have been proposed based on local self-similarity (LSS)
\cite{Schechtman07}, which is motivated by the notion that the
geometric layout of local internal self-similarities is relatively
insensitive to imaging properties. The state-of-the-art descriptor for
cross-modal dense correspondence, called dense adaptive
self-correlation (DASC) \cite{Kim15}, makes use of LSS and has
demonstrated high accuracy and speed on cross-modal image
pairs.
However, DASC suffers from two significant shortcomings.
One is its limited discriminative power due to
a limited set of patch sampling patterns used for modeling
internal self-similarities. In fact, the matching performance
of DASC may fall well short of CNN-based descriptors on images
that share the same modality.
The other major shortcoming is that the DASC descriptor does not provide
the flexibility to deal with non-rigid deformations, which
leads to lower robustness in matching.

In this paper, we introduce a novel descriptor, called deep
self-convolutional activations (DeSCA), that overcomes the
shortcomings of DASC while providing dense cross-modal
correspondences. This work is motivated by the observation that
local self-similarity can be formulated in a deep convolutional
architecture to enhance discriminative power and gain robustness to
non-rigid deformations. Unlike the DASC descriptor that selects
patch pairs within a support window and calculates the
self-similarity between them, we compute self-convolutional
activations that more comprehensively encode the intrinsic structure
by calculating the self-similarity between randomly selected patches
and all of the patches within the support window. These
self-convolutional responses are aggregated through spatial pyramid
pooling in a circular configuration, which yields a representation
less sensitive to non-rigid image deformations than the fixed patch
selection strategy used in DASC. To further enhance the
discriminative power and robustness, we build hierarchical self-convolutional
layers resembling a deep architecture used in CNN, together with
nonlinear and normalization layers. For efficient computation of
DeSCA over densely sampled pixels, we calculate the
self-convolutional activations through fast edge-aware filtering.

DeSCA resembles a CNN in its deep, multi-layer, and convolutional
structure. In contrast to existing CNN-based descriptors, 
DeSCA requires no training data for learning convolutional kernels,
since the convolutions are defined as the local self-similarity
between pairs of image patches, which yields its robustness to cross-modal imaging. 
\figref{img:1} illustrates the
robustness of DeSCA for image pairs across non-rigid deformations
and illumination changes. In the experimental results, we show that
DeSCA outperforms existing area-based and feature-based descriptors
on various benchmarks.\vspace{-5pt}

\section{Related Work}\label{sec:2}
\vspace{-5pt}
\subsubsection{Feature Descriptors}\label{sec:21}
Conventional gradient-based descriptors, such as SIFT \cite{Lowe04}
and DAISY \cite{Tola10}, as well as intensity comparison-based
binary descriptors, such as BRIEF \cite{Calonder11}, have shown
limited performance in dense correspondence estimation between
cross-modal image pairs. Besides these handcrafted features, several
attempts have been made using machine learning algorithms to derive
features from large-scale datasets \cite{Simonyan14,Trzcinski15}. A
few of these methods use deep convolutional neural networks (CNNs)
\cite{Alex12}, which have revolutionized image-level classification,
to learn discriminative descriptors for local patches. For designing
explicit feature descriptors based on a CNN architecture, immediate
activations are extracted as the descriptor
\cite{Simonyan14,Gong14,Fischer14,Donahue14,Simo-Serra15b}, and have been
shown to be effective for this patch-level task. However, even
though CNN-based descriptors encode a discriminative structure with
a deep architecture, they have inherent limitations in cross-modal
image
correspondence because %, like gradient-based descriptors,
they are derived from convolutional layers using shared patches or
volumes \cite{Simo-Serra15b,Dong15}. Furthermore, they cannot in
practice provide dense descriptors in the image domain due to their
prohibitively high computational complexity.

To estimate cross-modal correspondences, variants of the SIFT
descriptor have been developed \cite{Saleem14}, but these
gradient-based descriptors maintain an inherent limitation
similar to SIFT in dealing with image gradients that vary differently
between modalities. For illumination invariant correspondences, Wang
\emph{et al.} proposed the local intensity order pattern (LIOP)
descriptor \cite{Wang11}, but severe radiometric variations may
often alter the relative order of pixel intensities. Simo-Serra
\emph{et al.} proposed the deformation and light invariant (DaLI)
descriptor \cite{Simo-Serra15} to provide high resilience to
non-rigid image transformations and illumination changes, but it
cannot provide dense descriptors in the image domain due to its
high computational time.

Schechtman and Irani introduced the LSS descriptor
\cite{Schechtman07} for the purpose of template matching, and
achieved impressive results in object detection and retrieval. By
employing LSS, many approaches have tried to solve for cross-modal
correspondences \cite{Mattias12,Torabi13,Ye14}. However, none of
these approaches scale well to dense matching in cross-modal images
due to low discriminative power and high complexity. Inspired by
LSS, Kim \emph{et al.} recently proposed the DASC descriptor to
estimate cross-modal dense correspondences \cite{Kim15}. Though it
can provide satisfactory performance, it is not able to handle
non-rigid deformations and has limited discriminative power due to
its fixed patch pooling scheme. \vspace{-8pt}

\subsubsection{Area-Based Similarity Measures}\label{sec:22}
A popular measure for registration of cross-modal medical images is
mutual information (MI) \cite{Pluim03}, based on the entropy of the
joint probability distribution function, but it provides reliable
performance only for variations undergoing a global transformation
\cite{Heo13}. Although cross-correlation based methods such as
adaptive normalized cross-correlation (ANCC) \cite{Heo11} produce
satisfactory results for locally linear variations, they are less
effective against more substantial modality variations. Robust
selective normalized cross-correlation (RSNCC) \cite{Shen14} was
proposed for dense alignment between cross-modal images, but as an
intensity based measure it can still be sensitive to cross-modal
variations. Recently, DeepMatching \cite{Weinzaepfel13} was proposed to compute dense
correspondences by employing a hierarchical pooling scheme like CNN, but it is not designed to handle cross-modal
matching.\vspace{-5pt}
\begin{figure}[!t]
	\centering
	\renewcommand{\thesubfigure}{}
	\subfigure[(a)]
	{\includegraphics[width=0.2\linewidth]{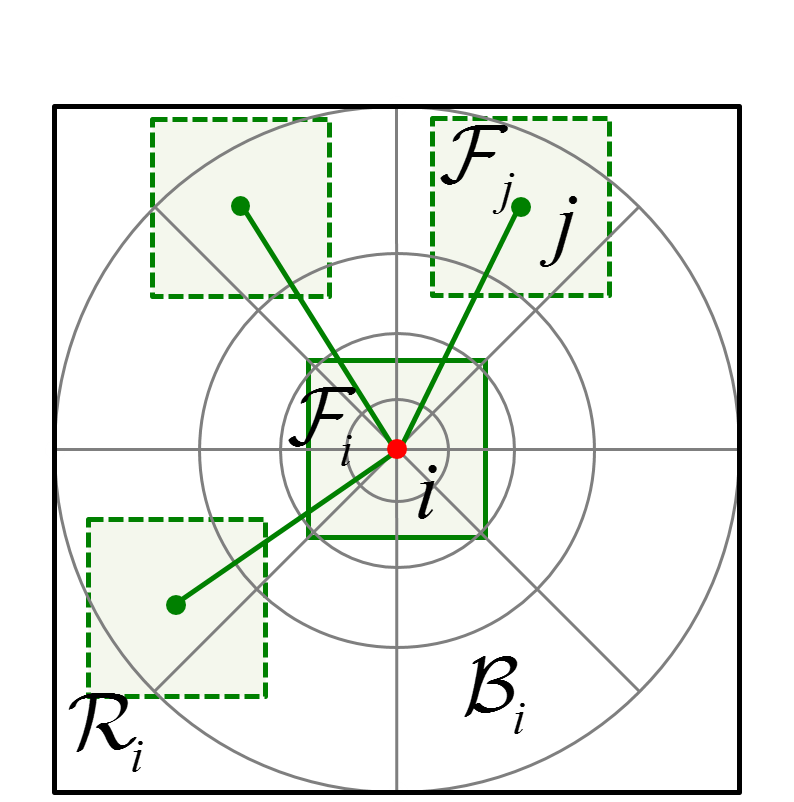}}\hfill
	\subfigure[(b)]
	{\includegraphics[width=0.2\linewidth]{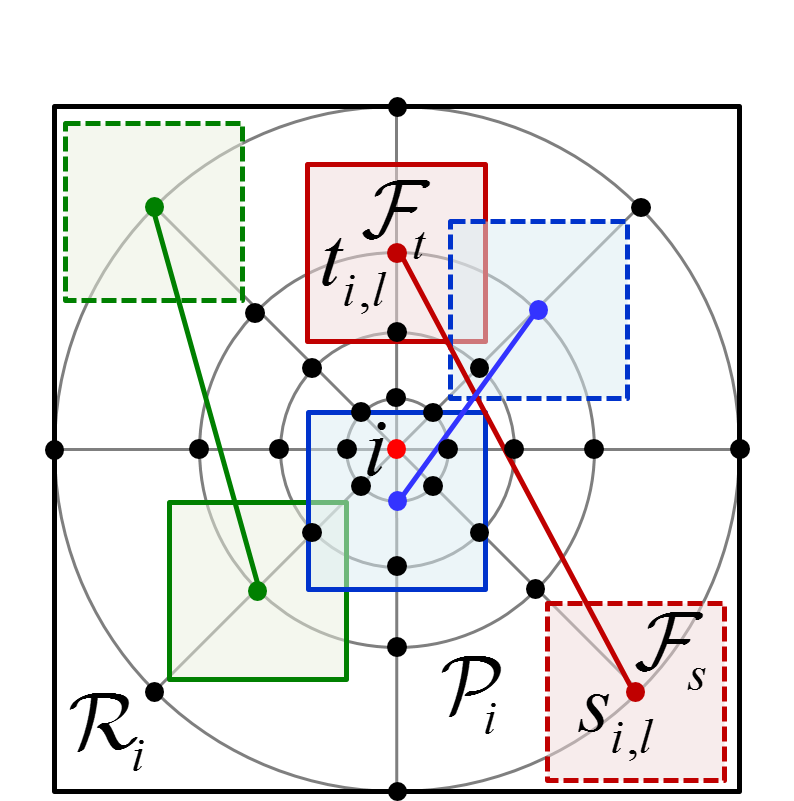}}\hfill
	\subfigure[(c)]
	{\includegraphics[width=0.6\linewidth]{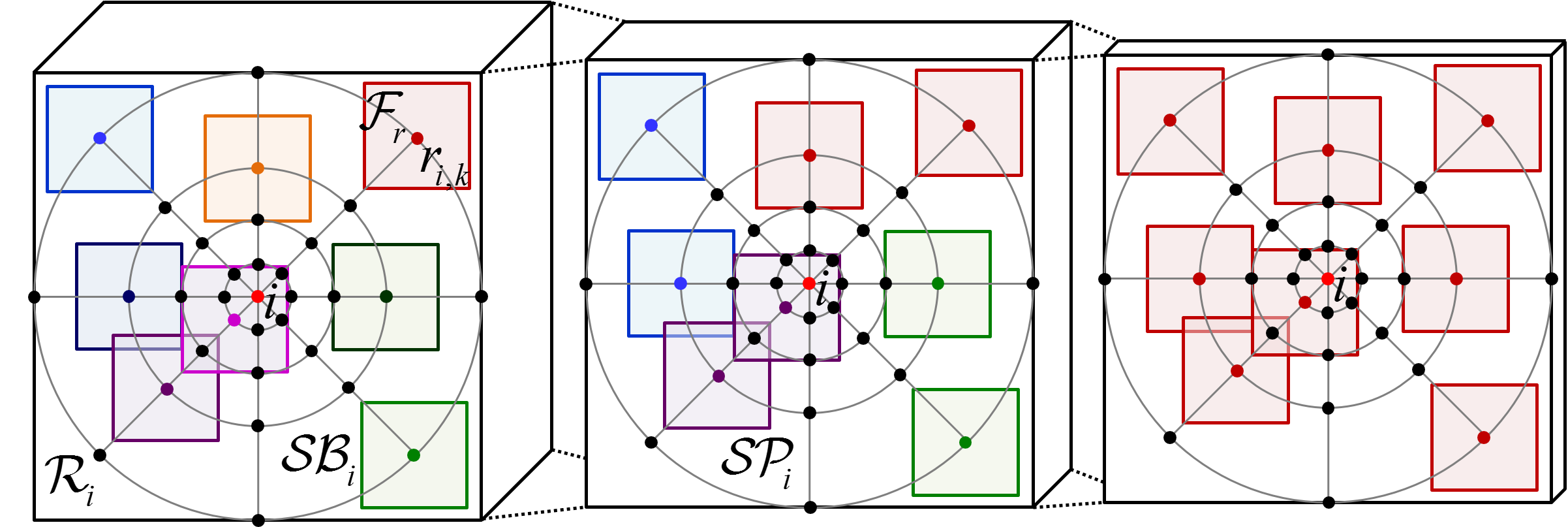}}\hfill
	\vspace{-12pt}
	\caption{Illustration of (a) LSS \cite{Schechtman07} using center-biased dense max pooling, (b) DASC \cite{Kim15} using patch-wise receptive field pooling, and (c) our DeSCA. Boxes, formed by solid and dotted lines, depict source and target patches. DeSCA incorporates a circular spatial pyramid pooling on hierarchical self-convolutional activations.
	}\label{img:2}\vspace{-10pt}
\end{figure}

\section{Background}\label{sec:3}
Let us define an image as ${f_i}:\mathcal{I} \to {\mathbb R}$ for pixel $i$, where
$\mathcal{I} \subset {{\mathbb N}^2}$ is a discrete image domain.
Given the image ${f_i}$, a dense descriptor ${\mathcal{D}_i}:\mathcal{I} \to \mathbb{R}^L$ with a feature dimension
of $L$ is defined on a local
support window $\mathcal{R}_i$ of size $M_\mathcal{R}$.

Unlike conventional descriptors, relying on common visual properties across images
such as color and gradient, LSS-based descriptors provide robustness to different imaging modalities
since internal self-similarities are preserved across cross-modal image pairs \cite{Schechtman07,Kim15}.
As shown in \figref{img:2}(a), the LSS discretizes
the correlation surface on a log-polar grid, generates a set of
bins, and then stores the maximum correlation value of each bin.
Formally, it generates an $L^{\textrm{LSS}}\times1$ feature vector $\mathcal{D}_{i}^{\textrm{LSS}} = { \bigcup
	_{l}}d_{i}^{\textrm{LSS}} (l)$ for $l \in \{1,...,L^{\textrm{LSS}}\}$,
with $d_{i}^{\textrm{LSS}} (l)$ computed as
\begin{equation}\label{equ:1}
d_{i}^{\textrm{LSS}} (l) = \mathop {\mathbf{max}}\limits_{j \in \mathcal{B}_{i}(l)}
\{ \mathbf{exp} (-\mathcal{S}(\mathcal{F}_i,\mathcal{F}_j)/\sigma_c) \},
\end{equation}
where log-polar bins are defined as $\mathcal{B}_{i} = \{j|j\in\mathcal{R}_i,\rho_{r-1}<{|i - j|}\leq\rho_{r},
\theta_{a-1}<{\angle (i - j)}\leq\theta_{a}\}$ with
a log radius $\rho_r$ for $r\in \{1,\cdots,N_\rho\}$ and a quantized
angle $\theta_a$ for $a\in \{1,\cdots,N_\theta\}$ with $\rho_{0}=0$
and $\theta_{0}=0$. $\mathcal{S}(\mathcal{F}_i,\mathcal{F}_j)$
is a correlation surface between a patch $\mathcal{F}_i$ and $\mathcal{F}_j$ of size $M_\mathcal{F}$, computed using sum of square differences.
Each pair of $r$ and $a$ is associated with a unique index $l$. Though LSS provides robustness to modality variations,
its significant computation does not scale well for estimating dense correspondences in cross-modal images.

Inspired by the LSS \cite{Schechtman07},
the DASC \cite{Kim15} encodes the similarity between patch-wise receptive
fields sampled from a log-polar circular point set $\mathcal{P}_{i}$ as
shown in \figref{img:2}(b).
It is defined such that $\mathcal{P}_{i} = \{j | j
\in \mathcal{R}_i, |{i} - {j}|=\rho_{r}, \angle ({i} -
{j})=\theta_{a} \}$, which has a higher density of points
near a center pixel, similar to DAISY \cite{Tola10}.
The DASC is encoded with a set of
similarities between patch pairs of sampling patterns
selected from $\mathcal{P}_{i}$ such that $\mathcal{D}^{\mathrm{DASC}}_{i} = {\bigcup_{l}}d^{\mathrm{DASC}}_{i} (l)$ for
$l \in \{1,...,L^{\textrm{DASC}}\}$:
\begin{equation}\label{equ:dasc}
\begin{array}{l}
d^{\mathrm{DASC}}_{i} (l) =
\mathbf{exp} ( - (1 - | {\mathcal{C} (\mathcal{F}_{s_{i,l}},\mathcal{F}_{t_{i,l}})} |)/\sigma_c ),\\
\end{array}
\end{equation}
where $s_{i,l}$ and $t_{i,l}$ are the $l^{th}$ selected sampling
pattern from $\mathcal{P}_{i}$ at pixel $i$. The patch-wise similarity is
computed with an exponential function with a bandwidth of
$\sigma_c$, which has been widely used for robust estimation
\cite{Black98}. ${\mathcal{C} (\mathcal{F}_{s_{i,l}},\mathcal{F}_{t_{i,l}})}$
is computed using an adaptive self-correlation measure. While the
DASC descriptor has shown satisfactory results for cross-modal dense
correspondence \cite{Kim15}, its randomized receptive field pooling
has limited descriptive power and does not accommodate non-rigid deformations.\vspace{-5pt}

\section{The DeSCA Descriptor}\label{sec:4}
\vspace{-5pt}
\subsection{Motivation and Overview}\label{sec:41}
Inspired by DASC \cite{Kim15}, our DeSCA descriptor also measures an adaptive self-correlation between two patches.
We, however, adopt a different strategy for selecting patch pairs,
and build self-convolutional activations that more comprehensively
encode self-similar structure to improve the
discriminative power and the robustness to non-rigid image
deformation (\secref{sec:42}). Motivated by the deep architecture of CNN-based
descriptors \cite{Simo-Serra15b}, we further build hierarchical
self-convolution activations to enhance the robustness of the DeSCA descriptor (\secref{sec:44}). Densely sampled descriptors are
efficiently computed over an entire image using a method based on
fast edge-aware filtering (\secref{sec:43}). \figref{img:2}(c)
illustrates the DeSCA descriptor,
which incorporates a circular spatial pyramid pooling
on hierarchical self-convolutional activations. \vspace{-5pt}
\begin{figure}[t]
	\centering
	\renewcommand{\thesubfigure}{}
	\subfigure[(a)]
	{\includegraphics[width=0.22\linewidth]{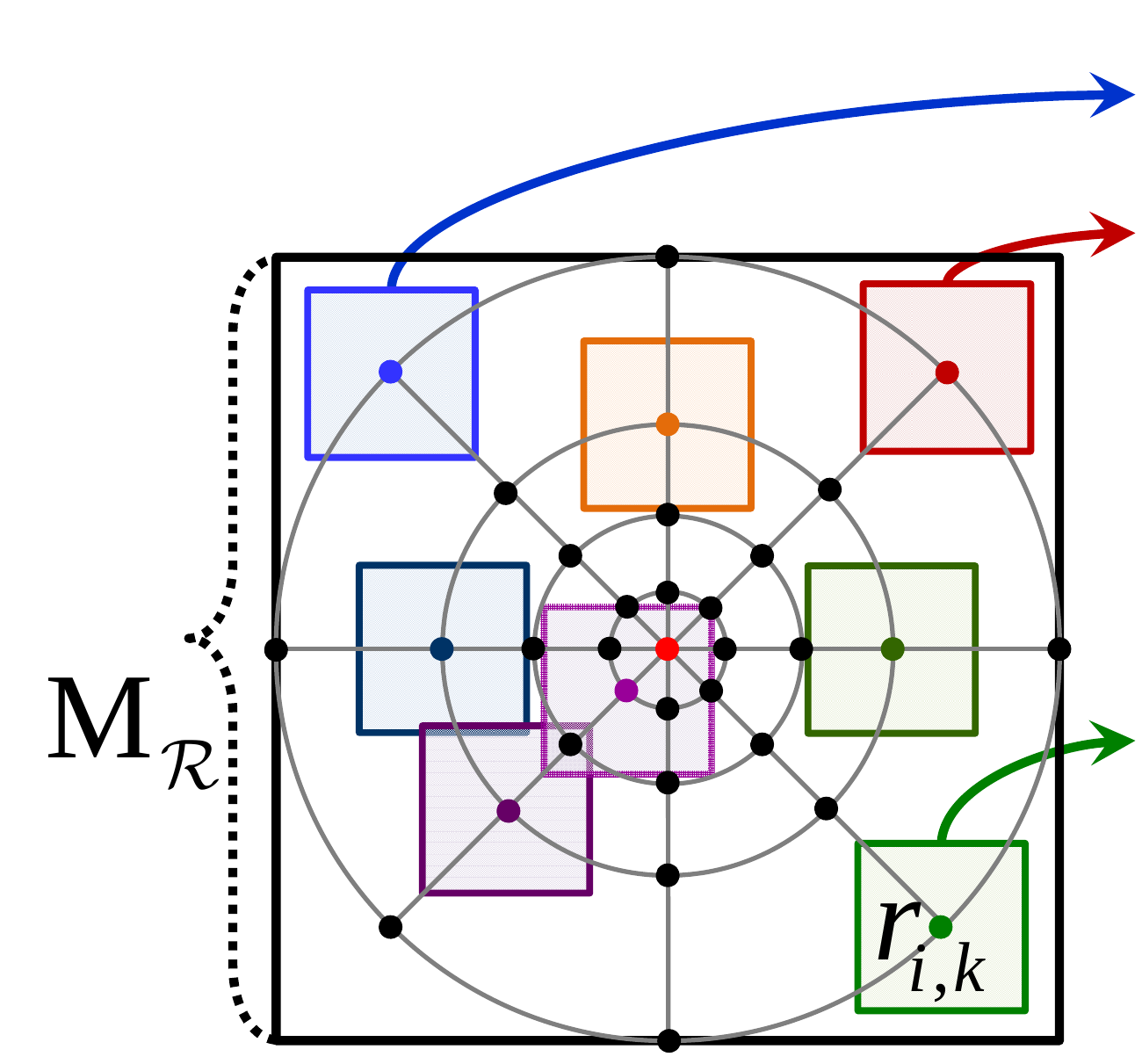}}
	\subfigure[(b)]
	{\includegraphics[width=0.22\linewidth]{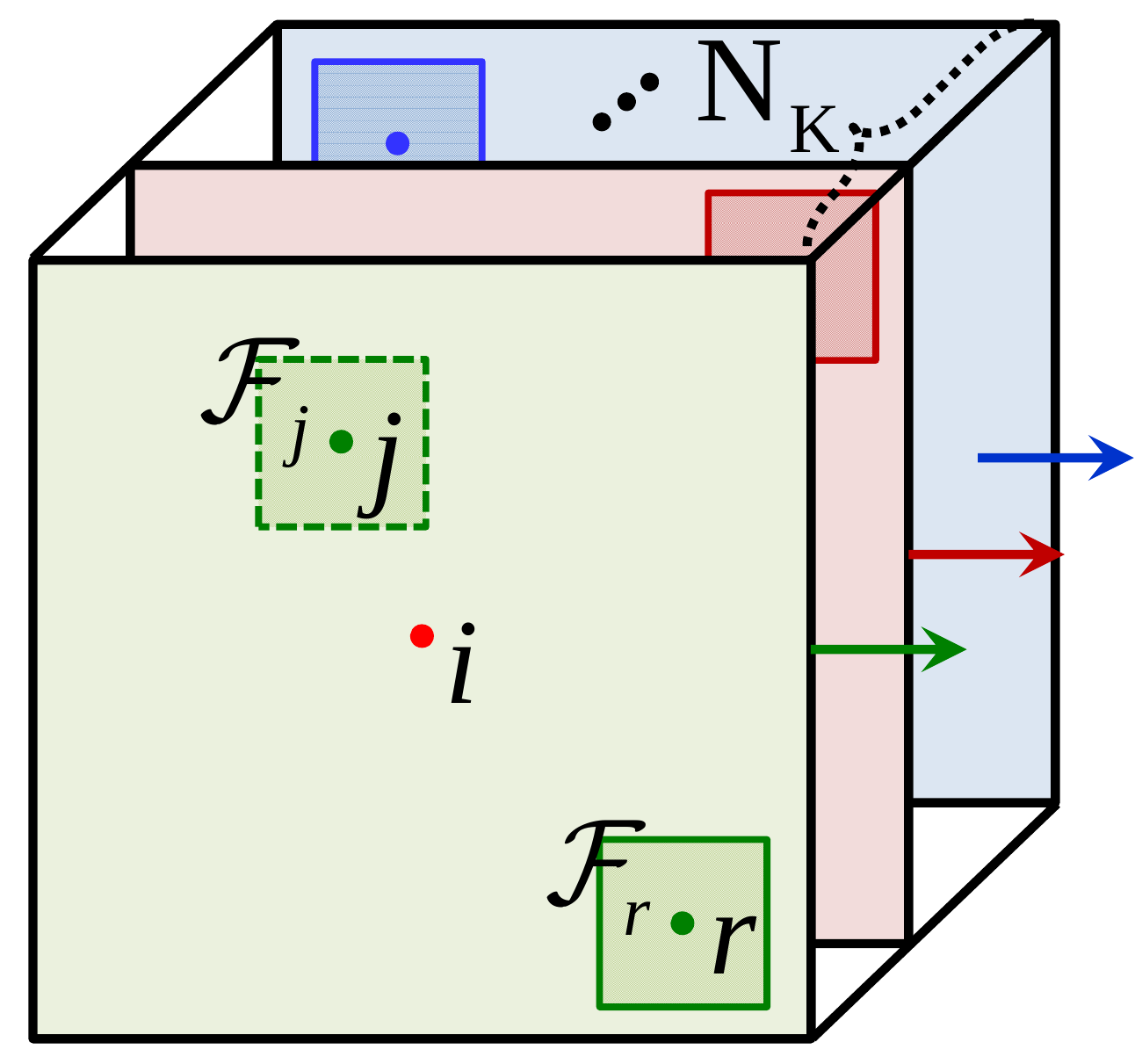}}
	\subfigure[(c)]
	{\includegraphics[width=0.22\linewidth]{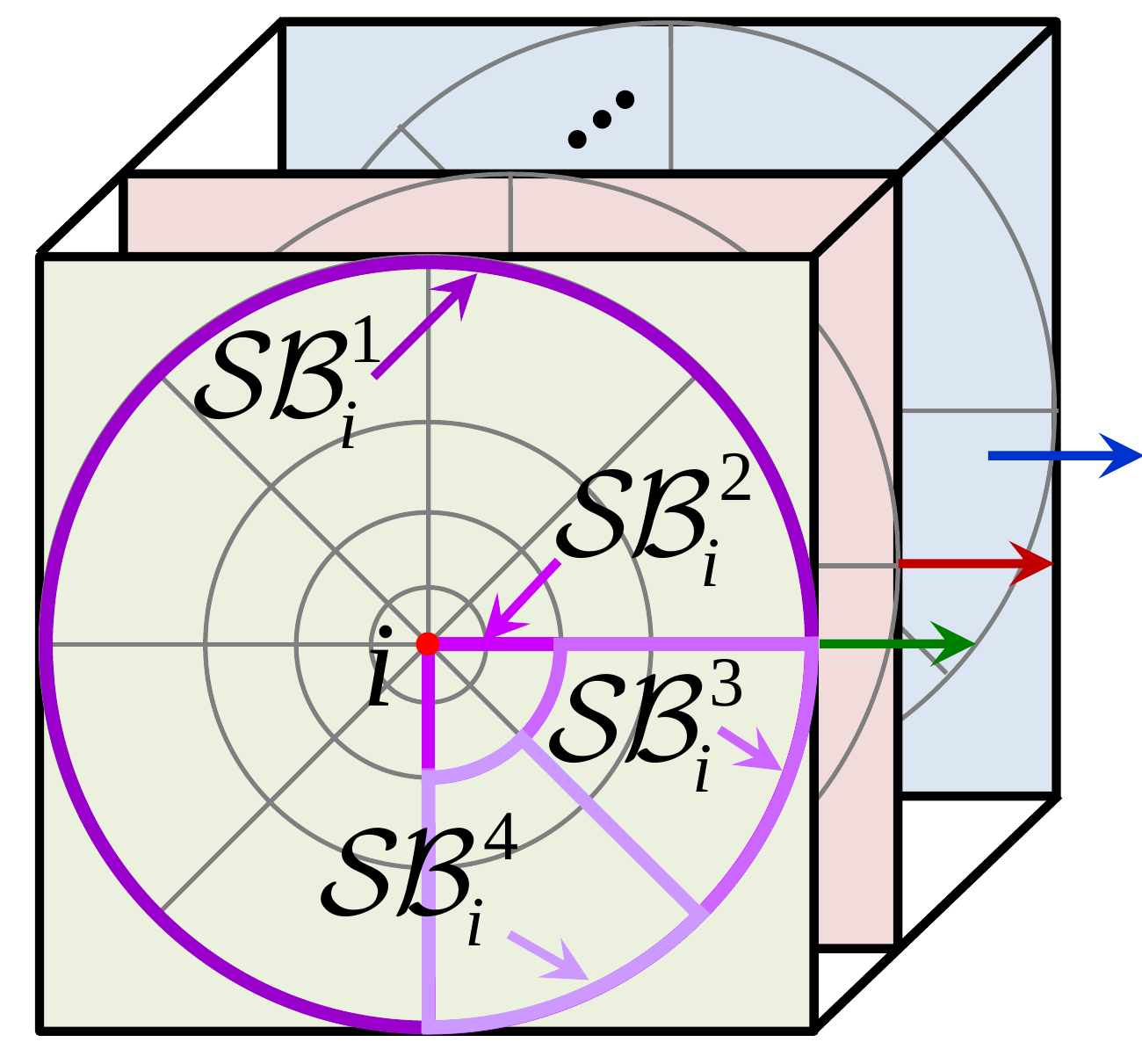}}
	\subfigure[(d)]
	{\includegraphics[width=0.22\linewidth]{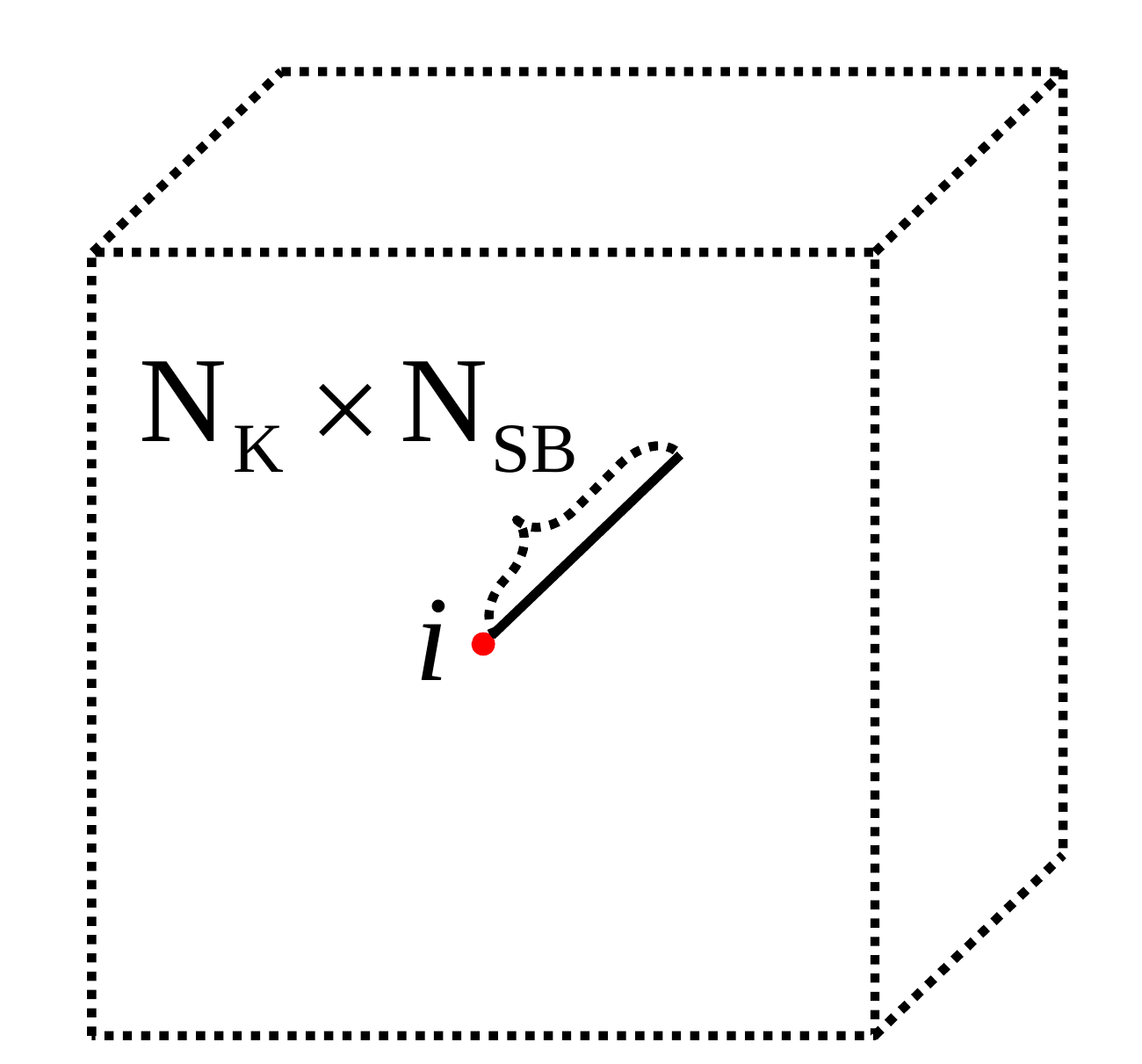}}\\
	\vspace{-12pt}
	\caption{Computation of single self-convolutional activation (SiSCA).
		(a) A local support window $\mathcal{R}_i$ of size $M_\mathcal{R}^2$ with $N_K$ random
		samples. (b) For each random patch, a self-convolutional surface is computed using an adaptive self-correlation measure.
		(c) A self-convolutional activation is then obtained through circular spatial pyramid pooling (C-SPP).
		(d) The activation from C-SPP is concatenated as 1-D feature vector.}\label{img:3}\vspace{-5pt}
\end{figure}
\begin{figure}[!t]
	\centering
	\renewcommand{\thesubfigure}{}
	\subfigure[(a) $s=1$]
	{\includegraphics[width=0.12\linewidth]{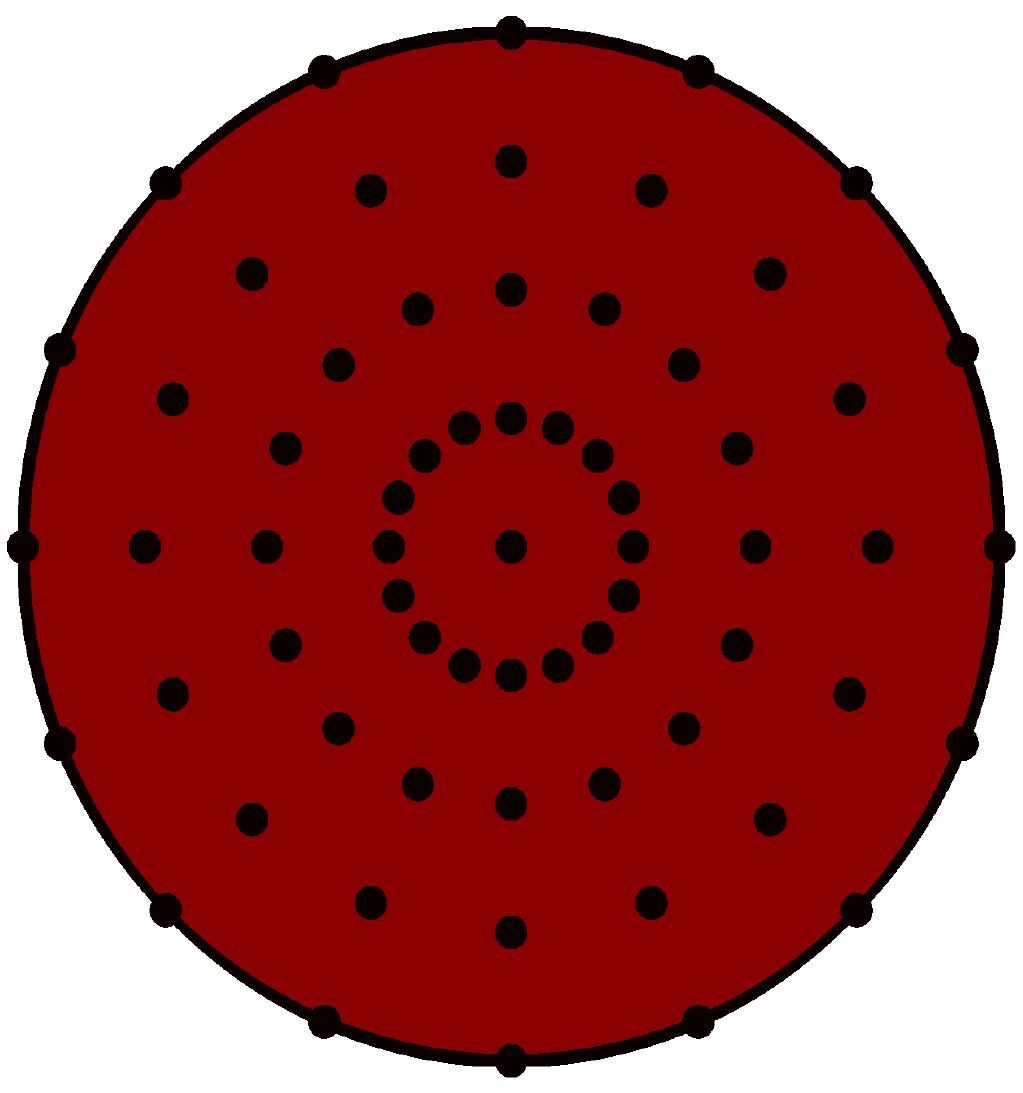}}
	\subfigure[(b) $s=2$]
	{\includegraphics[width=0.12\linewidth]{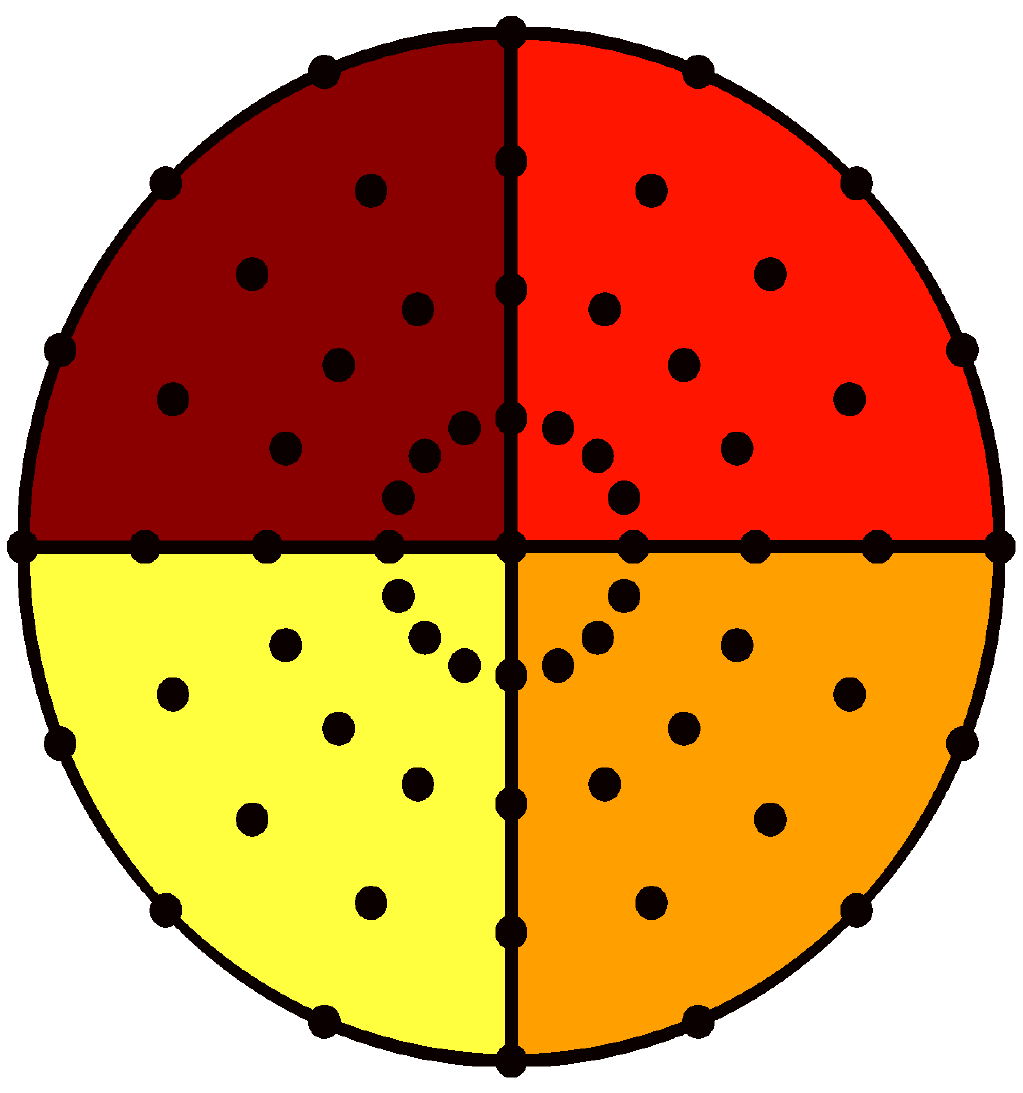}}
	\subfigure[(c) $s=3$]
	{\includegraphics[width=0.12\linewidth]{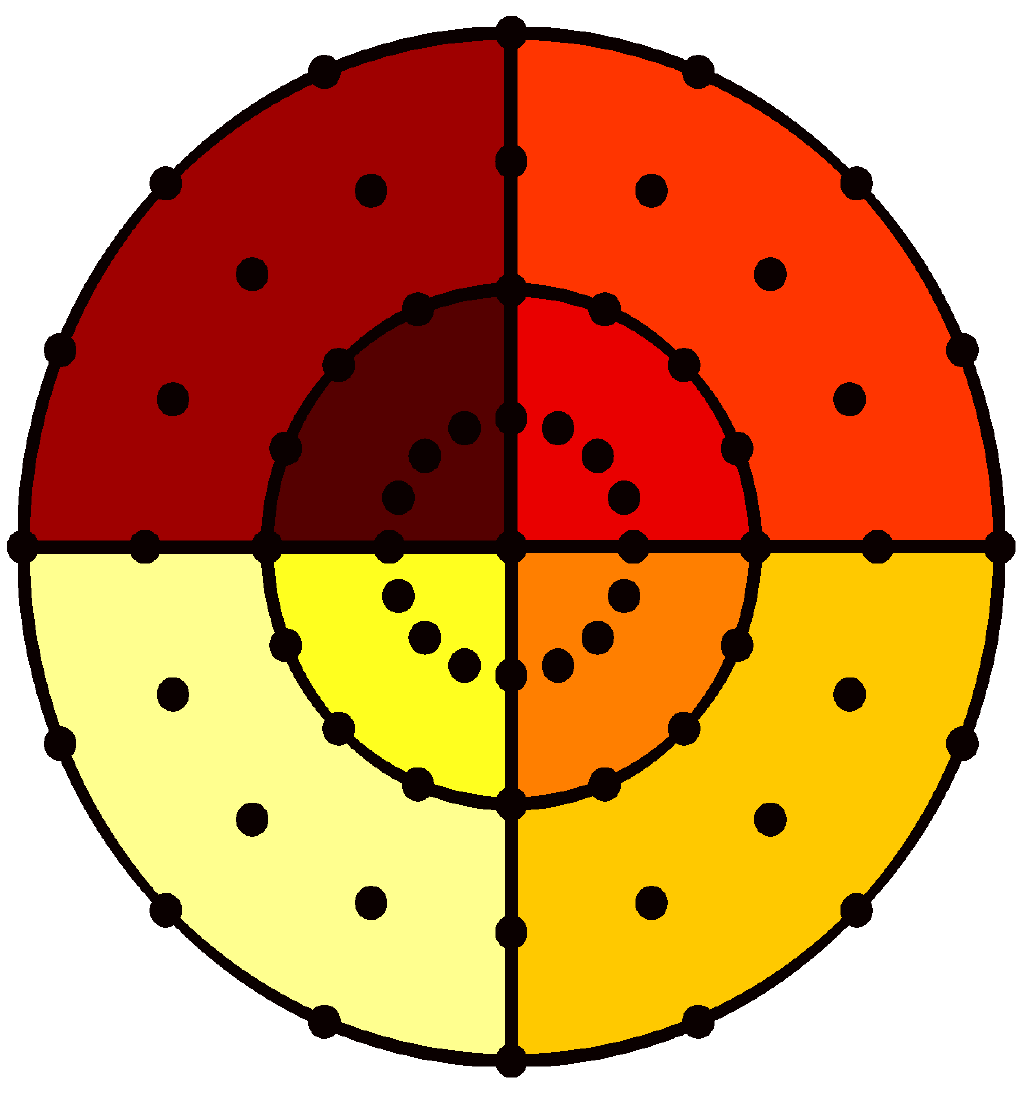}}
	\subfigure[(b) $s=4$]
	{\includegraphics[width=0.12\linewidth]{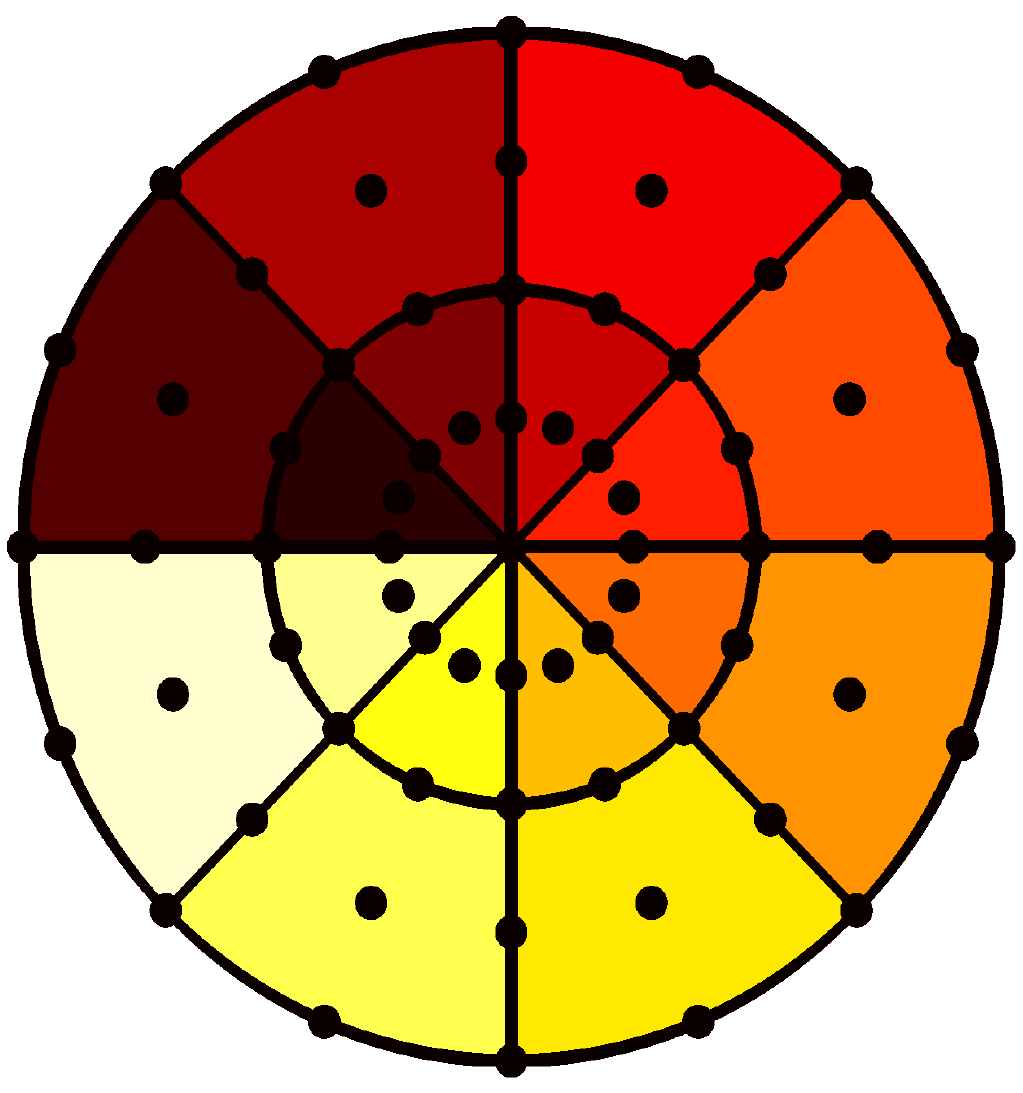}}
	\subfigure[(e) $s=5$]
	{\includegraphics[width=0.12\linewidth]{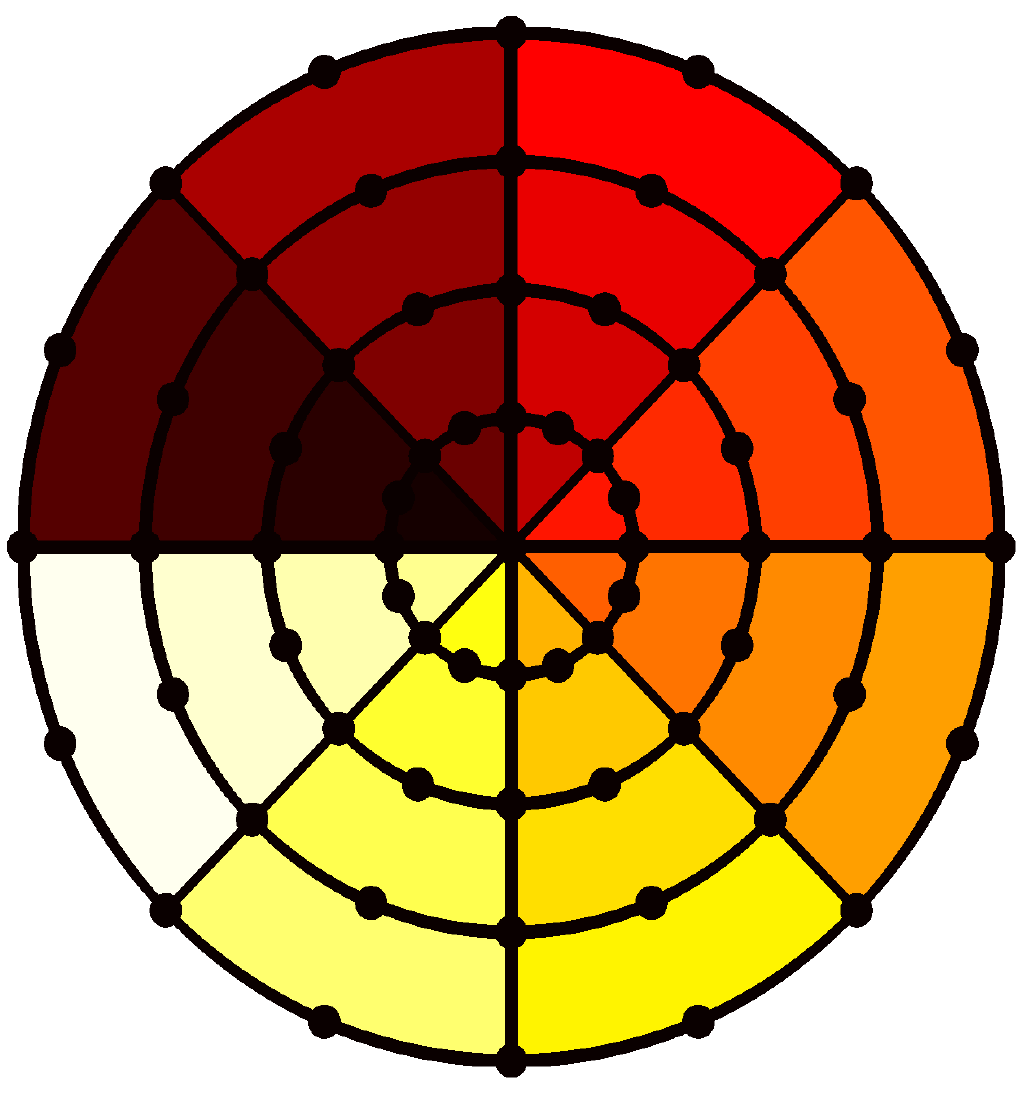}}\\
	\vspace{-12pt}
	\caption{Examples of the circular spatial pyramidal bins $\mathcal{SB}_{i}$.
		The total number of bins is $N_{\mathcal{SB}} = \sum^{N_S}\nolimits_{s=2} 2^s + 1$,
		where $N_S$ represents the pyramid level.}\label{img:4}\vspace{-10pt}
\end{figure}

\subsection{SiSCA: Single Self-Convolutional Activation}\label{sec:42}
To simultaneously leverage the benefits of self-similarity in DASC \cite{Kim15}
and the deep convolutional archiecture of CNNs while overcoming the limitations of each method,
our approach builds self-convolutional activations.
Unlike DASC \cite{Kim15}, the feature response is obtained 
through circular spatial pyramid pooling.
We start by describing a single-layer version of DeSCA,
which we denote as SiSCA.\vspace{-8pt}

\subsubsection{Self-Convolutions}\label{sec:421}
To build a self-convolutional activation, we randomly select $N_K$
points from a log-polar circular point set $\mathcal{P}_{i}$ defined
within a local support window $\mathcal{R}_i$. We convolve a patch
$\mathcal{F}_{r_{i,k}}$ centered at the $k$-th point ${r_{i,k}}$
with all patches $\mathcal{F}_j$, which is defined for $j \in
\mathcal{R}_i$ and $k \in \{1,...,N_K\}$ as \figref{img:3}(b). 
Similar to DASC \cite{Kim15}, the similarity
$\mathcal{C}(\mathcal{F}_{r_{i,k}},\mathcal{F}_j)$ between patch
pairs is measured using an adaptive self-correlation, which is
known to be effective in addressing cross-modality. With $(i,k)$
omitted for simplicity, $\mathcal{C}(\mathcal{F}_r,\mathcal{F}_j)$
is computed as follows:
\begin{equation}\label{equ:asasc}
\mathcal{C}(\mathcal{F}_r,\mathcal{F}_j) = \frac{\sum\nolimits_{r',j'}
	{\omega _{r,r'} ({f_{r'}} - {\mathcal{G}_{r,r}}) ({f_{j'}} - {\mathcal{G}_{r,j}})}} {\sqrt{\sum\nolimits_{r'} {\omega_{r,r'}}({f_{r'}} - {\mathcal{G}_{r,r}}) }
	\sqrt{\sum\nolimits_{r',j'} {\omega_{r,r'}({f_{j'}} - {\mathcal{G}_{r,j}})}}},
\end{equation}
for $r' \in \mathcal{F}_{r}$ and $j' \in\mathcal{F}_{j}$.
${\mathcal{G}_{r,r}}=\sum\nolimits_{r'} {{\omega_{r,r'}}{f_{r'}}}$
and ${\mathcal{G}_{r,j}}=\sum\nolimits_{r',j'}{{\omega
		_{r,r'}}{f_{j'}}}$ represent weighted averages of $f_{r'} \in \mathcal{F}_r$
and $f_{j'} \in \mathcal{F}_j$.
Similar to DASC \cite{Kim15}, the weight ${\omega _{r,r'}}$
represents how similar two pixels $r$ and $r'$ are, and is
normalized, \emph{i.e.}, $\sum\nolimits_{r'} {{\omega _{r,r'}}}=1$.
It may be defined using any form of edge-aware weighting
\cite{Gastal11,He13}.\vspace{-8pt}

\subsubsection{Circular Spatial Pyramid Pooling}\label{sec:422}
To encode the feature responses on the self-convolutional surface,
we propose a circular spatial pyramid pooling (C-SPP) scheme, which
pools the responses within each hierarchical spatial bin, similar to
a spatial pyramid pooling (SPP) \cite{Dong15,Seidenari14,He15} but in
a circular configuration. Note that many existing descriptors also
adopt a circular pooling scheme thanks to its robustness based on a
higher pixel density near a central pixel
\cite{Schechtman07,Tola10,Calonder11}. We further encodes more
structure information with a C-SPP.
\begin{figure}[!t]
	\centering
	\renewcommand{\thesubfigure}{}
	\subfigure[(a)]
	{\includegraphics[width=0.22\linewidth]{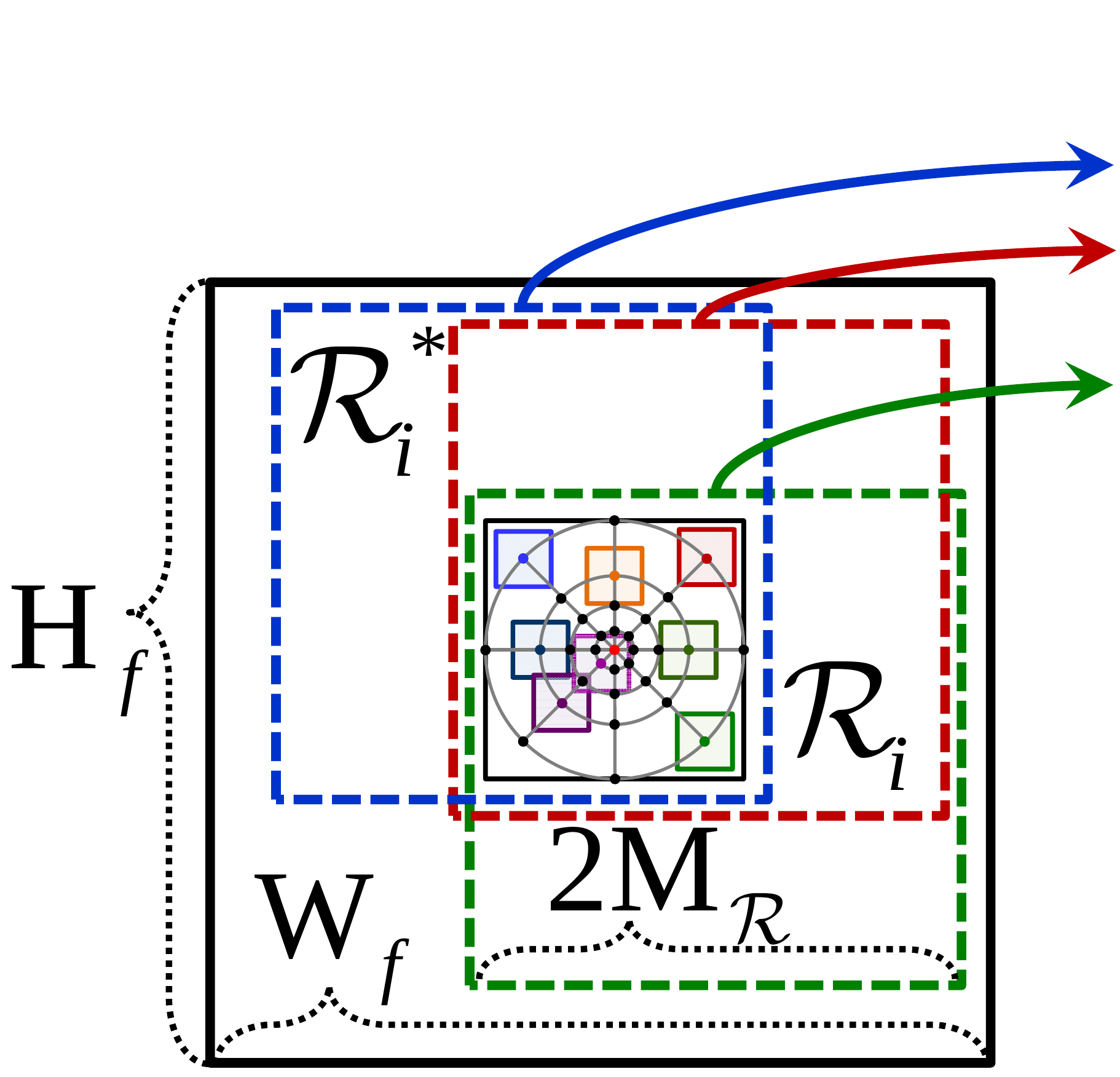}}
	\subfigure[(b)]
	{\includegraphics[width=0.22\linewidth]{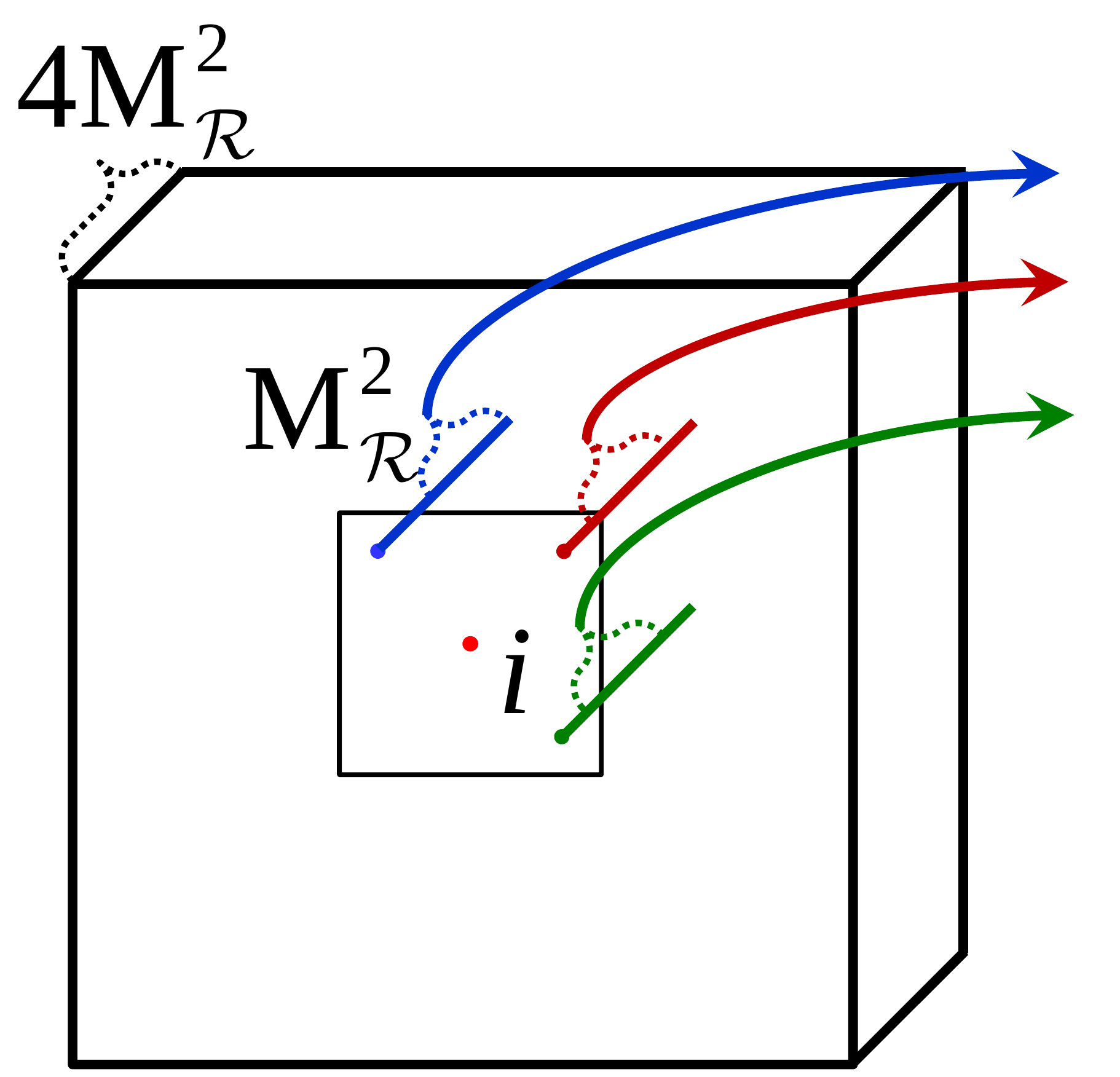}}
	\subfigure[(c)]
	{\includegraphics[width=0.22\linewidth]{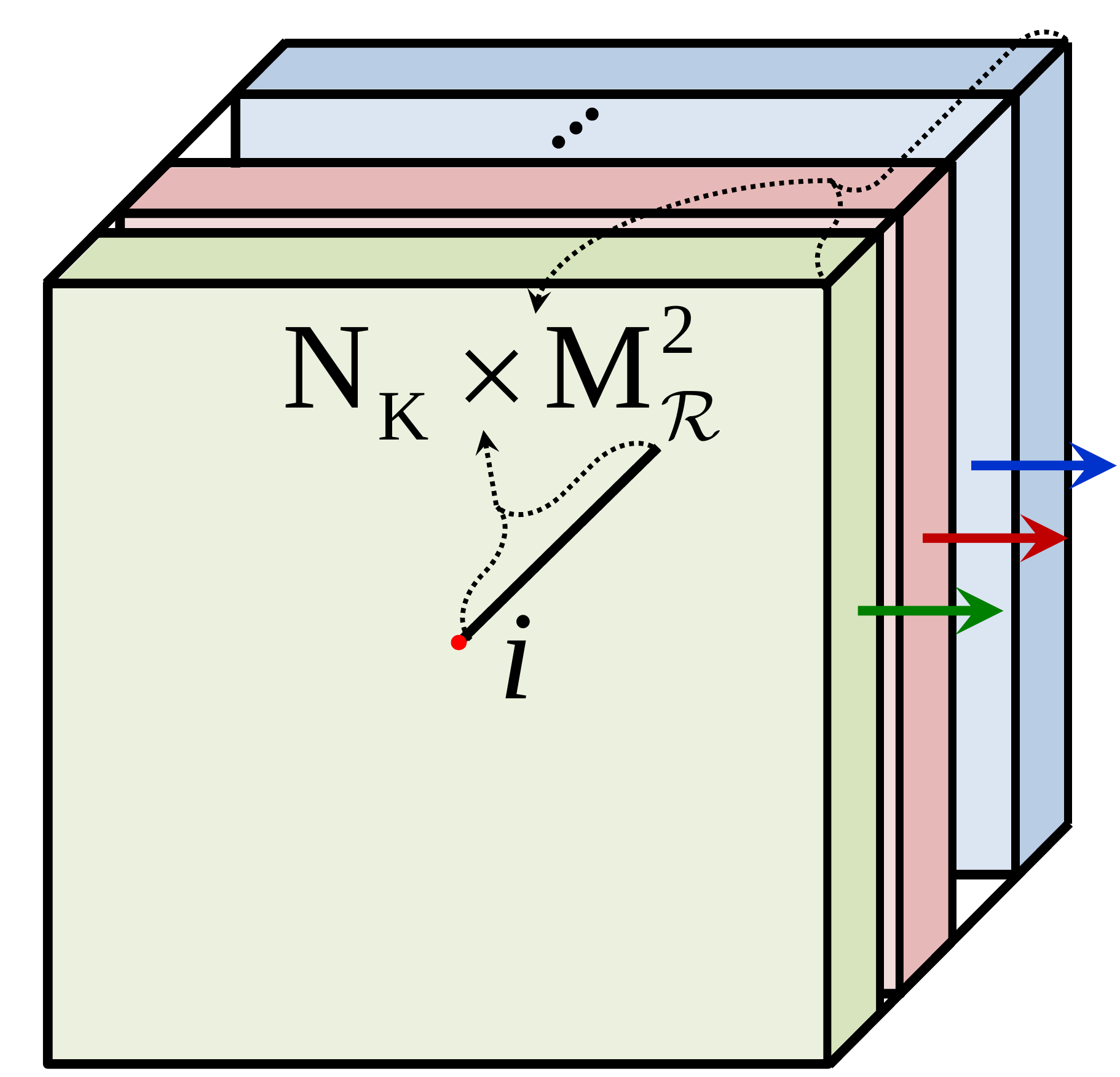}}
	\subfigure[(d)]
	{\includegraphics[width=0.22\linewidth]{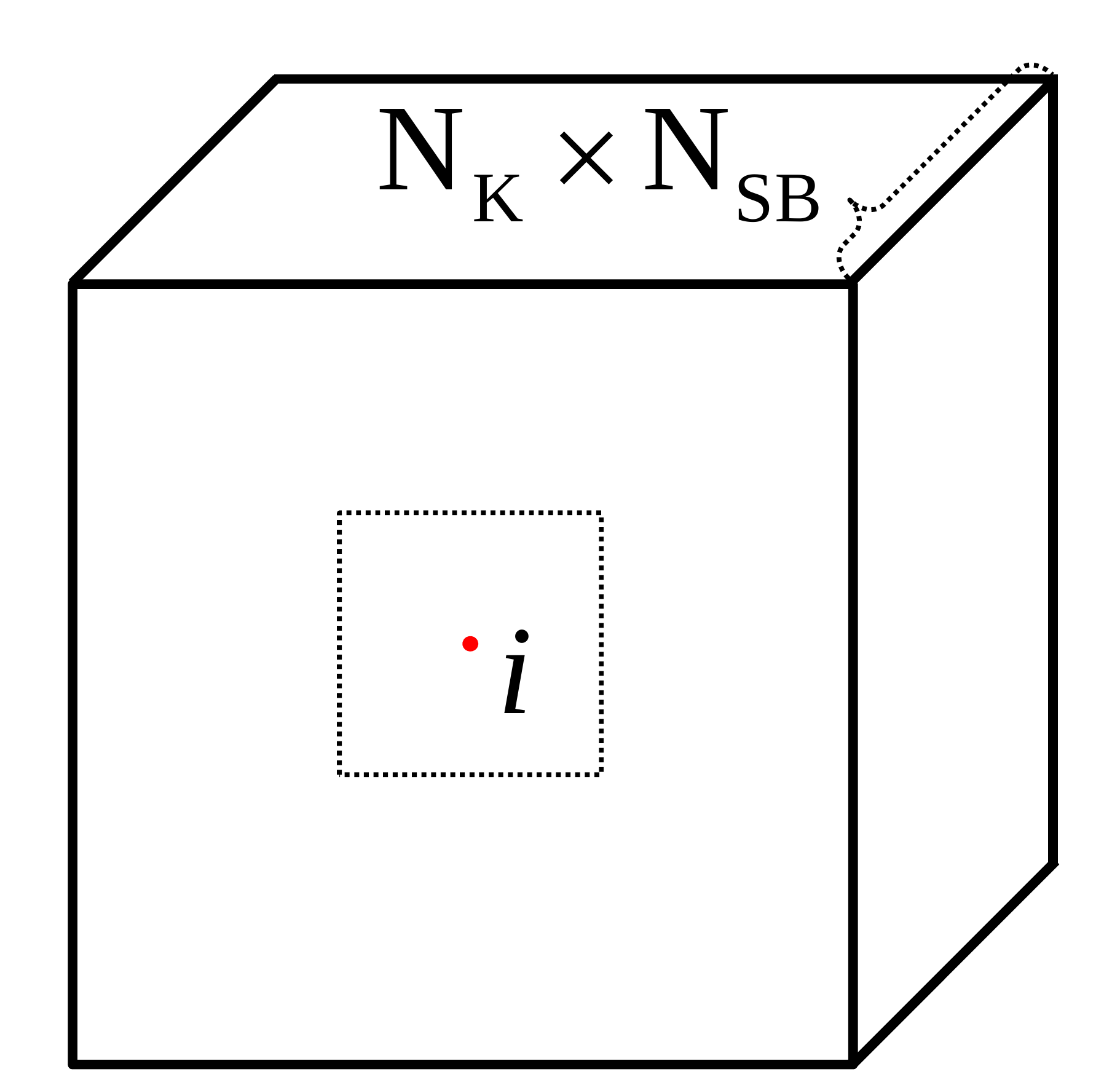}}\\
	\vspace{-12pt}
	\caption{Efficient computation of self-convolutional activations on the image. (a) An image $f_i$ with
		a doubled support window $\mathcal{R}^*_i$ and random samples. (b) 1-D vectorial self-convolutional surface.
		(c) Self-convolutional activations. (d) Activations after C-SPP.
		With an efficient edge-aware filtering and activation reformulation,
		self-convolutonal activations are computed efficiently in a dense manner.}\label{img:5}\vspace{-10pt}
\end{figure}

The circular pyramidal bins $\mathcal{SB}_{i}(u)$ are
defined from log-polar circular bins $\mathcal{B}_{i}$, where $u$
indexes all pyramidal level $s \in \{1,...,N_S\}$ and all bins in
each level $s$ as in \figref{img:4}. The circular pyramidal bin at the
top of pyramid, \emph{i.e.}, $s=1$, first encompasses all of bins
$\mathcal{B}_{i}$. At the second level, \emph{i.e.}, $s=2$, it is
defined by dividing $\mathcal{B}_{i}$ into quadrants. For further
lower pyramid levels, \emph{i.e.}, $s>2$, the circular pyramidal
bins are defined differently according to whether $s$ is odd or
even. For an odd $s$, the bins are defined by dividing bins in upper
level into two parts along the radius. For an even $s$, they are
defined by dividing bins in upper level into two parts with respect
to the angle. The set of all circular pyramidal bins
$\mathcal{SB}_{i}$ is denoted such that $\mathcal{SB}_{i} =
\bigcup\nolimits_{u} \mathcal{SB}_{i} (u)$ for $u \in
\{1,...,N_{\mathcal{SB}}\}$, where the number of circular spatial
pyramid bins is defined as $N_{\mathcal{SB}} =
\sum^{N_S}\nolimits_{s=2} 2^s + 1$.

As illustrated in \figref{img:3}(c), the feature responses are
finally max-pooled on the circular pyramidal bins
$\mathcal{SB}_{i}(u)$ of each self-convolutional surface
$\mathcal{C}(\mathcal{F}_{r_{i,k}},\mathcal{F}_j)$, yielding a
feature response
\begin{equation}\label{equ:scpp}
h_i (k,u) = \mathop {\mathbf{max}}\limits_{j \in
	\mathcal{SB}_{i}(u)} \{
\mathcal{C}(\mathcal{F}_{r_{i,k}},\mathcal{F}_j) \},
\quad u \in \{1,...,N_\mathcal{SB}\}.
\end{equation}

This pooling is repeated for all $k \in \{1,...,N_K\}$, yielding 
accumulated activations $\hat{h}_i (l) = {\bigcup\nolimits_{\{k,u\}}{h_i (k,u)}}$ 
where $l$ indexes for all $k$ and $u$.

Interestingly, LSS \cite{Schechtman07} also uses
the max pooling strategy to mitigate the effects of non-rigid image
deformation. However, max pooling in the 2-D self-correlation surface of LSS \cite{Schechtman07} loses 
fine-scale matching details as reported in \cite{Kim15}. By
contrast, DeSCA employs circular spatial pyramid
pooling in the 3-D self-correlation surface that provides a more
discriminative representation of self-similarities, thus maintaining
fine-scale matching details as well as providing robustness
to non-rigid image deformations.
\vspace{-8pt}

\subsubsection{Non-linear Gating and Nomalization}\label{sec:422}
The final feature responses are passed through a non-linear and
normalization layer to mitigate the effects of outliers. With
accumulated activations $\hat{h}_i$, the single self-convolution
activiation (SiSCA) descriptor $\mathcal{D}^{\mathrm{SiSCA}}_{i} =
{\bigcup _{l}}d^{\mathrm{SiSCA}}_{i} (l)$ is computed for $l \in
\{1,...,L^{\mathrm{SiSCA}}\}$ through a non-linear gating layer:
\begin{equation}\label{equ:sisca_surf}
d^{\mathrm{SiSCA}}_{i} (l)
= \mathbf{exp} ( - (1 - | \hat{h}_i (l) |)/\sigma_c ),
\end{equation}
where $\sigma_c$ is a Gaussian kernel bandwidth.
The size of features obtained from the SiSCA
becomes $L^{\mathrm{SiSCA}}=N_K N_{\mathcal{SB}}$.
Finally, $d^{\mathrm{SiSCA}}_{i} (l)$ for each pixel $i$
is normalized with an L-2 norm for all $l$.\vspace{-5pt}
\begin{figure}[!t]
	\centering
	\renewcommand{\thesubfigure}{}
	\subfigure[]
	{\includegraphics[width=0.8\linewidth]{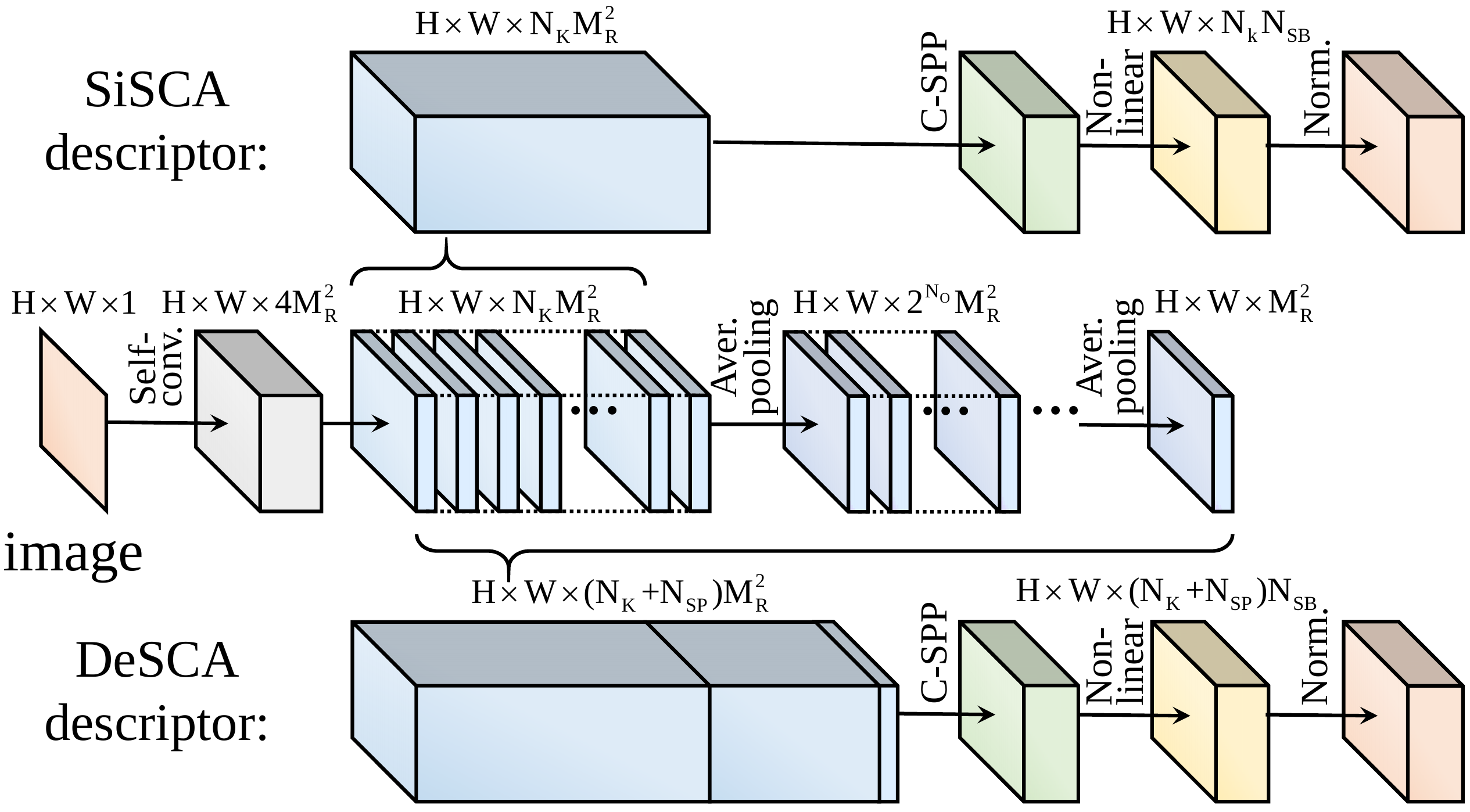}}
	\vspace{-15pt}
	\caption{Visualization of SiSCA and DeSCA descriptor.
		Our architecture consists of a hierarchical self-convolutional layer,
		circular spatial pyramid pooling layer, non-linear gating layer, and normalization layer.}\label{img:6}\vspace{-10pt}
\end{figure}

\subsection{Efficient Computation for Dense Description}\label{sec:43}
The most time-consuming part of DeSCA is in constructing
self-convolutional surfaces
$\mathcal{C}(\mathcal{F}_{r_{i,k}},\mathcal{F}_j)$ for $k$ and $j$,
where $N_K M^2_\mathcal{R}$ computations of \eqref{equ:asasc} are
needed for each pixel $i$. Straightforward computation of a weighted
summation using $\omega$ in \equref{equ:asasc} would require
considerable processing with a computational complexity of $O(I
M_{\mathcal{F}} N_K M^2_\mathcal{R})$, where $I = H_f W_f$ represents the image size (height $H_f$ and width
$W_f$). To expedite processing, we
utilize fast edge-aware filtering \cite{Gastal11,He13} and propose a
pre-computation scheme for convolutional surfaces.

Similar to DASC \cite{Kim15}, we compute $\mathcal{C}(\mathcal{F}_{r_{i,k}},\mathcal{F}_j)$ efficiently
by first rearranging the sampling patterns $(r_{i,k},j)$ into
reference-biased pairs $(i,j_r) = (i,i+r_{i,k}-j)$.
$\mathcal{C}(\mathcal{F}_i,\mathcal{F}_{j_r})$ can then be expressed as
\begin{equation}\label{equ:asasc_app}
\mathcal{C}(\mathcal{F}_i,\mathcal{F}_{j_r}) =
\frac{{{\mathcal{G}_{i,ij_r}}  - {\mathcal{G}_{i,i}}  \cdot
		{\mathcal{G}_{i,j_r}} }} {{\sqrt {{\mathcal{G}_{i,i^{2}}} -
			{(\mathcal{G}_{i,i})^2}} \cdot \sqrt {{\mathcal{G}_{i,j^{2}_r}}  -
			{{(\mathcal{G}_{i,j_r})^2}}} }},
\end{equation}
where ${\mathcal{G}_{i,ij_r}}=\sum\nolimits_{i',j'_r}{{\omega
		_{i,i'}}{f_{i'}}{f_{j'_r}}}$,
${\mathcal{G}_{i,j_r^{2}}}=\sum\nolimits_{i',j'_r}
{{\omega_{i,i'}}{f_{j'_r}^{2}}}$, and ${\mathcal{G}_{i,i^{2}}} =
\sum\nolimits_{i'} {{\omega_{i,i'}}f_{i'}^2} $.
$\mathcal{C}(\mathcal{F}_i,\mathcal{F}_{j_r})$ can be efficiently
computed using any form of fast edge-aware filter
\cite{Gastal11,He13} with the complexity of $O(I N_K M^2_\mathcal{R})$.
$\mathcal{C}(\mathcal{F}_{r_{i,k}},\mathcal{F}_j)$ is then simply
obtained from $\mathcal{C}(\mathcal{F}_i,\mathcal{F}_{j_r})$ by
re-indexing sampling patterns.

Though we remove the computational dependency on patch size
$M_\mathcal{F}$, $N_K M^2_\mathcal{R}$ computations of
\eqref{equ:asasc_app} are still needed to obtain the
self-convolutional activations, where many sampling pairs are
repeated. To avoid such redundancy, we first compute
self-convolutional activation
$\mathcal{C}(\mathcal{F}_i,\mathcal{F}_j)$ for $j \in
\mathcal{R}^*_i$ with a doubled local support window
$\mathcal{R}^*_i$ of size $2M_\mathcal{R} \times 2M_\mathcal{R} (= 4
M_\mathcal{R}^2)$. A doubled local support window is used because
\eqref{equ:asasc_app} is computed with patch $\mathcal{F}_{j_r}$ and
the minimum support window size for $\mathcal{R}^*_i$ to cover all
samples within $\mathcal{R}_i$ is $2M_\mathcal{R}$ as shown in
\figref{img:5}(b). After the self-convolutional activation for
$\mathcal{R}^*_i$ is computed once over the image domain,
$\mathcal{C}(\mathcal{F}_{r_{i,k}},\mathcal{F}_j)$ can be extracted 
through an index mapping process, where the indexes for
$\mathcal{R}_{i-r_{i,k}}$ are estimated from
$\mathcal{R}^*_i$.\vspace{-5pt}
\begin{table}[!t]
	\begin{center}
		\begin{tabularx}{\linewidth}{p{4mm} p{1mm}| p{160mm}}
			\hlinewd{0.8pt}
			\multicolumn{3}{ p{165mm} }{{\bf Algorithm 1}: Deep Self-Convolutional Activations (DeSCA) Descriptor}\\
			\hlinewd{0.8pt}
			\multicolumn{3}{ p{165mm} }{{\bf Input} : image ${f_i}$, random samples ${r_{i,k}}$.}\\
			\multicolumn{3}{ p{165mm} }{{\bf Output} : DeSCA descriptor ${\mathcal{D}^{\mathrm{DeSCA}}_{i}}$.}\\
			$\mathbf{1:}$&\multicolumn{2}{ p{165mm} }{Compute $\mathcal{C}(\mathcal{F}_i,\mathcal{F}_j)$ for a doubled support window $\mathcal{R}^*_i$ by using \equref{equ:asasc_app}.}\\
			$\mathbf{2:}$&\multicolumn{2}{ p{165mm} }{Estimate $\mathcal{C}(\mathcal{F}_{r_{i,k}},\mathcal{F}_j)$ from $\mathcal{C}(\mathcal{F}_i,\mathcal{F}_j)$ according to the index mapping process.}\\
			~&\multicolumn{2}{ p{165mm} }{{\bf for } $v = 1:N_{\mathcal{SP}}$ {\bf do } {\bf $/*$ hierarchical aggregation using average pooling $*/$}}\\
			$\mathbf{3:}$&~&$\;\;$Determine a circular pyramidal point $\mathcal{SP}_{i}(v)$.\\
			$\mathbf{4:}$&~&$\;\;$Compute $\mathcal{C}(\mathcal{F}_{v},\mathcal{F}_j)$ by using an average pooling for $\mathcal{SP}_{i}(v)$ on $\mathcal{C}(\mathcal{F}_{r_{i,k}},\mathcal{F}_j)$.\\
			~&\multicolumn{2}{ p{165mm} }{{\bf end for }}\\
			~&\multicolumn{2}{ p{165mm} }{{\bf for } $u = 1:N_{\mathcal{SB}}$ {\bf do } {\bf $/*$ hierarchical spatial aggregation using C-SPP $*/$}}\\
			$\mathbf{6:}$&~&$\;\;$Determine a circular pyramidal bin $\mathcal{SB}_{i}(u)$.\\
			$\mathbf{7:}$&~&$\;\;$Compute $h_i (k,u)$ and $h_i (v,u)$ by using C-SPP on each $\mathcal{SB}_{i}(u)$ \par$\;\;$from $\mathcal{C}(\mathcal{F}_{r_{i,k}},\mathcal{F}_j)$ and $\mathcal{C}(\mathcal{F}_{v},\mathcal{F}_j)$, respectively.\\
			~&\multicolumn{2}{ p{165mm} }{{\bf end for }}\\
			$\mathbf{8:}$&\multicolumn{2}{ p{165mm} }{Build hierarchical self-convolutional activations $\hat{h}_i (l)$ from $h_i (k,u)$ and $h_i (v,u)$.}\\
			$\mathbf{8:}$&\multicolumn{2}{ p{165mm} }{Compute a nonlinear response \equref{equ:sisca_surf}, followed by L-2 normalization.}\\
			$\mathbf{9:}$&\multicolumn{2}{ p{165mm} }{Build a DeSCA descriptor $\mathcal{D}^{\mathrm{DeSCA}}_{i} = {\bigcup _{l}}d^{\mathrm{DeSCA}}_{i} (l)$.}\\
			\hlinewd{0.8pt}
		\end{tabularx}
	\end{center}\label{alg:1}\vspace{-20pt}
\end{table}

\subsection{DeSCA: Deep Self-Convolutional Activations}\label{sec:44}
So far, we have discussed how to build the self-convolutional
activation on a single level. In this section, we extend this idea
by encoding self-similar structures at multiple levels in a manner
similar to a deep architecture widely adopted in the CNNs
\cite{Alex12}. DeSCA is defined similarly to SiSCA, except that an
average pooling is executed before C-SPP (see \figref{img:6}). 
With self-convolutional activations, we perform the average pooling on
circular pyramidal point sets.

In comparison to the self-convolutions just from a
single patch, the spatial aggregation of self-convolutional responses is
clearly more robust, and it requires only marginal computational overhead over SiSCA. 
The strength of such a hierarchical
aggregation has also been shown in \cite{Weinzaepfel13}. Compared to 
using only last CNN layer activations,
we use all intermediate activations from hierarchical average
pooling, which yields better cross-modal matching quality.
\begin{figure}[!t]
	\centering
	\renewcommand{\thesubfigure}{}
	\subfigure[(a)]
	{\includegraphics[width=0.25\linewidth]{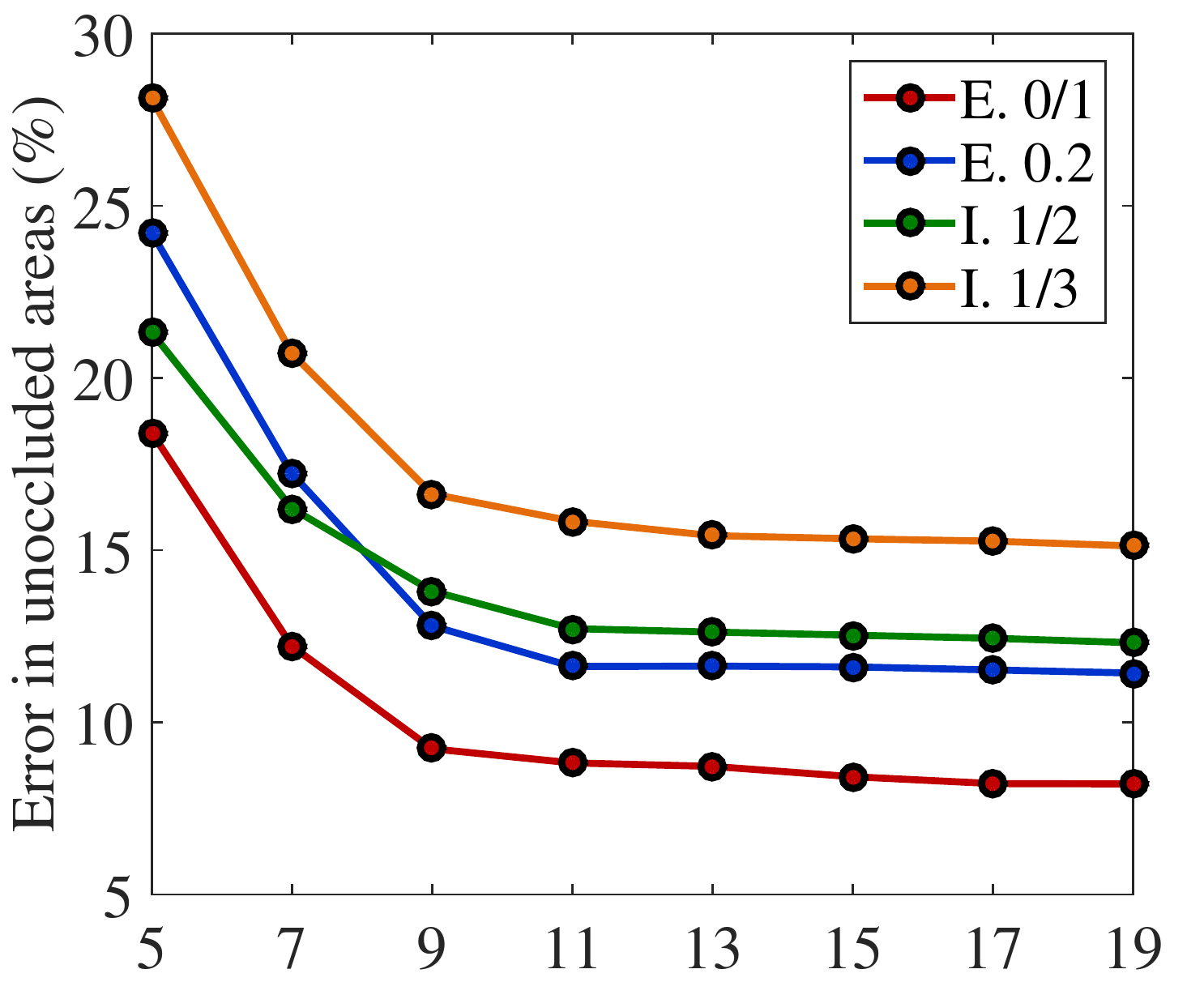}}\hfill
	\subfigure[(b)]
	{\includegraphics[width=0.25\linewidth]{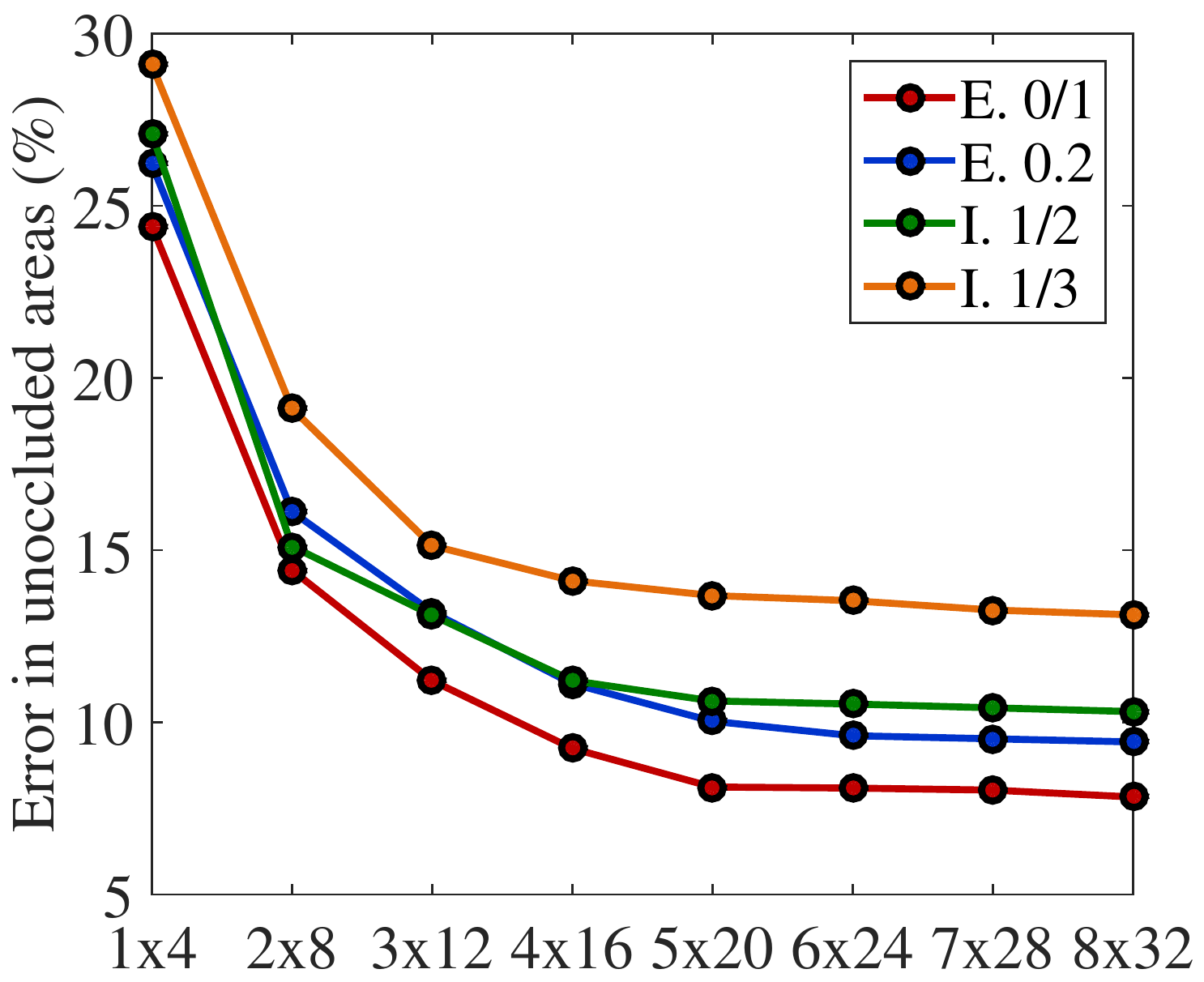}}\hfill
	\subfigure[(c)]
	{\includegraphics[width=0.25\linewidth]{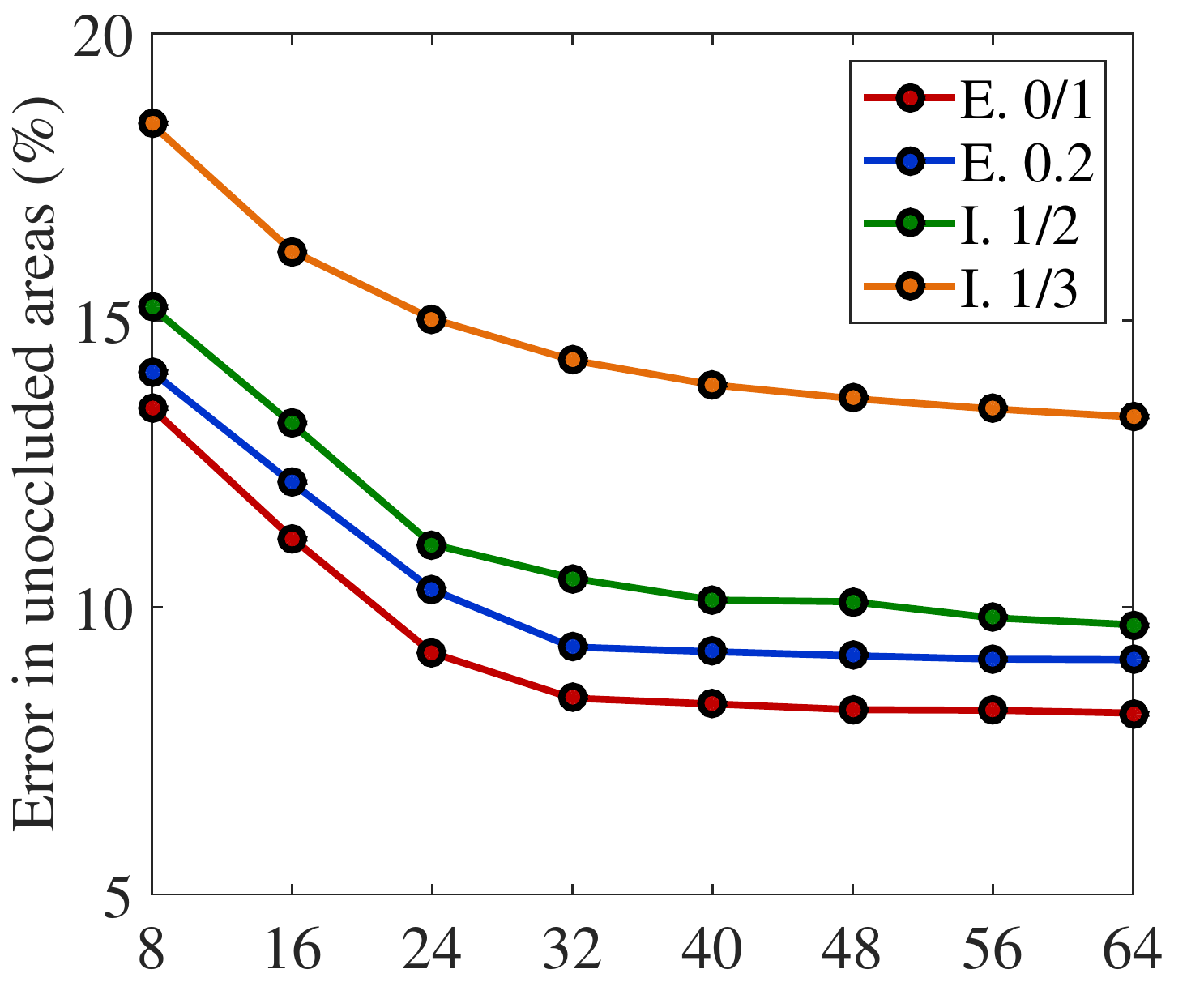}}\hfill
	\subfigure[(d)]
	{\includegraphics[width=0.25\linewidth]{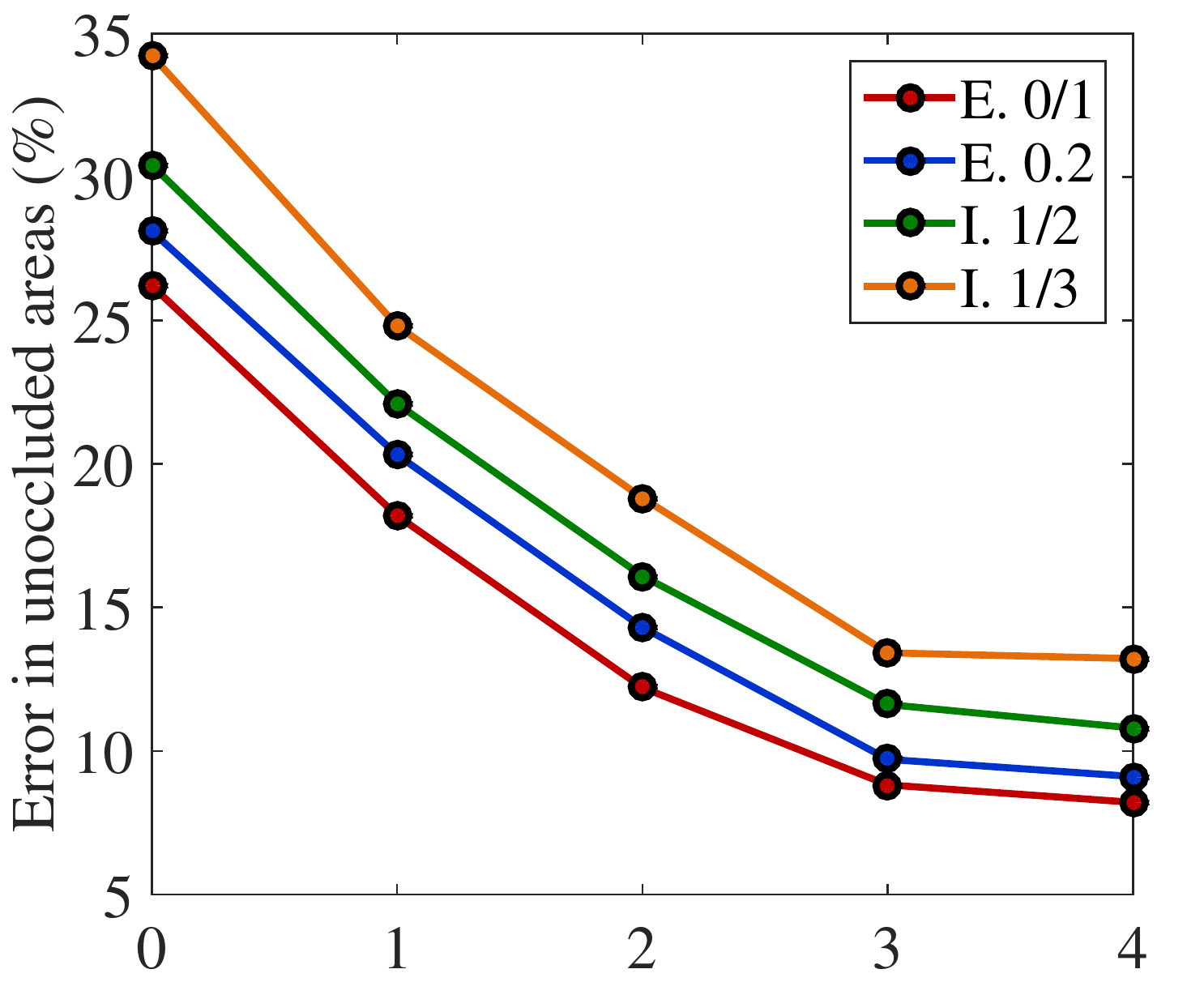}}\hfill
	\vspace{-12pt}
	\caption{Component analysis of DeSCA on the Middlebury benchmark \cite{middlebury}
		for varying parameter values, such as (a) support window size $M_\mathcal{R}$, (b) number of log-polar circular point $N_\rho \times N_\theta$,
		(c) number of random samples $N_K$, and (d) level of circular spatial pyramid $N_S$.
		In each experiment, all other parameters are fixed to the initial values.}\label{img:7}\vspace{-10pt}
\end{figure}

To build the hierarchical self-convolutional volume using an average pooling,
we first define the circular pyramidal point sets $\mathcal{SP}_{i}(v)$
from log-polar circular point sets $\mathcal{P}_{i}$,
where $v$ associates all pyramidal level $o \in \{1,...,N_O\}$ and
all points in each level $o$. In the average pooling, 
the circular pyramidal bins $\mathcal{SB}_{i}(u)$ used in C-SPP
is re-used such that $\mathcal{SP}_{i}(v) = 
\{ j | j \in \mathcal{P}_{i}, j \in \mathcal{SB}_{i}(u)\}$, thus $N_S = N_O$. 
Deep self-convolutional activations are defined by aggregating $\mathcal{C}(\mathcal{F}_{r_{i,k}},\mathcal{F}_j)$ for
all $r_{i,k}$ patches determined on each $\mathcal{SP}_{i}(v)$ such that
\begin{equation}\label{equ:deepscs}
\mathcal{C}(\mathcal{F}_{v},\mathcal{F}_j) = \sum\nolimits_{r_{i,k} \in \mathcal{SP}_{i}(v)} \mathcal{C}(\mathcal{F}_{r_{i,k}},\mathcal{F}_j) / N_{v},
\end{equation}
which is defined for all $v$, and $N_{v}$ is the number of $r_{i,k}$
patches within $\mathcal{SP}_{i}(v)$. The hierarchical activations
are sequentially aggregated using average pooling from bottom to top
of circular pyramidal point set $\mathcal{SP}_{i}(v)$. 
After computing hierarchical self-convolutional
aggregations, similar to SiSCA, the DeSCA employs C-SPP, non-linear,
and normalization layer presented in \secref{sec:42}. Hierarchical
self-convolutional activation ${h_i (v,u)}$ is computed using the
C-SPP such that
\begin{equation}\label{equ:scpp_deep}
h_i (v,u) = \mathop {\mathbf{max}}\limits_{j \in \mathcal{SB}_{i}(u)} \{ \mathcal{C}(\mathcal{F}_{v},\mathcal{F}_j) \}.
\end{equation}

Accumulated self-convolutional activations are built from $h_i (k,u)$ in \equref{equ:scpp} and 
$h_i (v,u)$ in \equref{equ:scpp_deep} such that $\hat{h}_i (l) = {\bigcup\nolimits_{\{k,v,u\}}{\{h_i (k,u),h_i (v,u)\}}}$.
Our DeSCA descriptor $d^{\mathrm{DeSCA}}_{i} (l)$ is then passed 
through a non-linear layer. 
$\mathcal{D}^{\mathrm{DeSCA}}_{i} = {\bigcup _{l}}d^{\mathrm{DeSCA}}_{i} (l)$ 
is built for $l \in \{1,...,L^{\mathrm{DeSCA}}\}$ with 
$L^{\mathrm{DeSCA}} = (N_K+N_{\mathcal{SP}}) N_{\mathcal{SB}}$.
Finally, $d^{\mathrm{DeSCA}}_{i} (l)$ for each pixel $i$
is normalized with an L-2 norm for all $l$.\vspace{-5pt}

\section{Experimental Results and Discussion}\label{sec:5}
\vspace{-5pt}
\subsection{Experimental Settings}\label{sec:51}
In our experiments, the DeSCA descriptor was implemented with the following fixed parameter settings
for all datasets: $\{\sigma_c,M_\mathcal{F},M_\mathcal{R},N_K,N_S\}  = \{ 0.5,5,9,32,3\}$,
and $\{N_\rho,N_\theta\}  = \{4,16\}$.
We chose the guided filter (GF) for edge-aware filtering in \equref{equ:asasc_app},
with a smoothness parameter of $\epsilon=0.03^2$.
We implemented the DeSCA descriptor in C++ on an Intel Core i7-3770 CPU at 3.40 GHz.
We will make our code publicly available.
The DeSCA descriptor was compared to other state-of-the-art
descriptors (SIFT \cite{Lowe04}, DAISY \cite{Tola10}, BRIEF
\cite{Calonder11}, LIOP \cite{Wang11}, DaLI \cite{Simo-Serra15}, LSS \cite{Schechtman07}, and DASC \cite{Kim15}),
as well as area-based approaches (ANCC \cite{Heo11} and RSNCC \cite{Shen14}).
Furthermore, to evaluate the performance gain with a deep architecture,
we compared SiSCA and DeSCA.\vspace{-5pt}
\begin{figure}[!t]
	\centering
	\renewcommand{\thesubfigure}{}
	\subfigure[]
	{\includegraphics[width=0.140\linewidth]{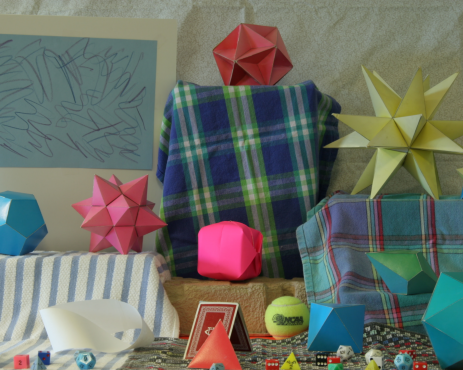}}\hfill
	\subfigure[]
	{\includegraphics[width=0.140\linewidth]{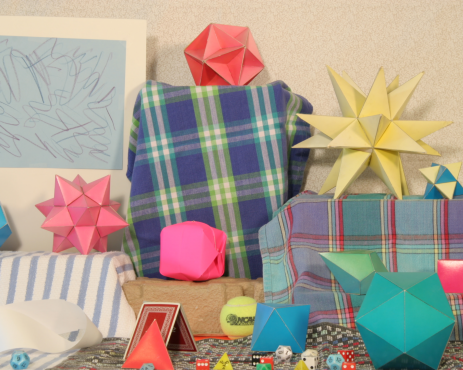}}\hfill
	\subfigure[]
	{\includegraphics[width=0.140\linewidth]{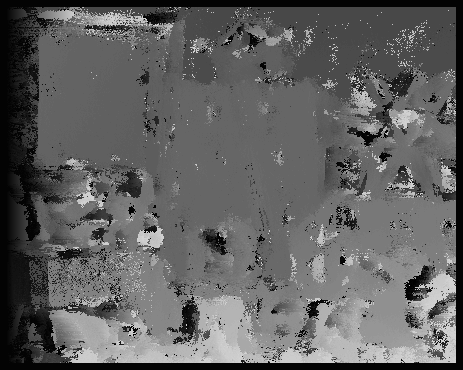}}\hfill
	\subfigure[]
	{\includegraphics[width=0.140\linewidth]{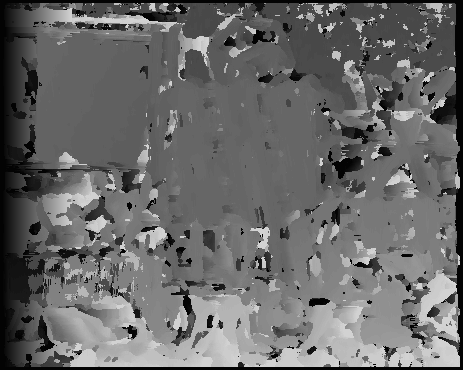}}\hfill
	\subfigure[]
	{\includegraphics[width=0.140\linewidth]{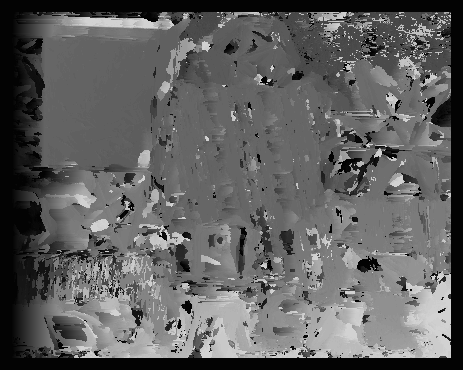}}\hfill
	\subfigure[]
	{\includegraphics[width=0.140\linewidth]{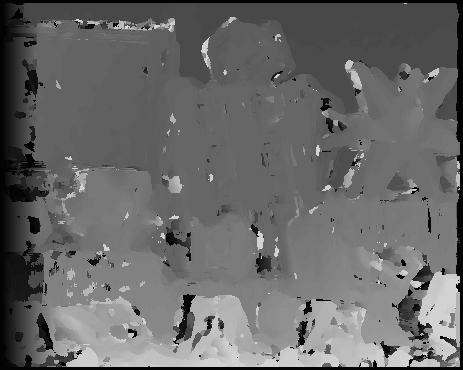}}\hfill
	\subfigure[]
	{\includegraphics[width=0.140\linewidth]{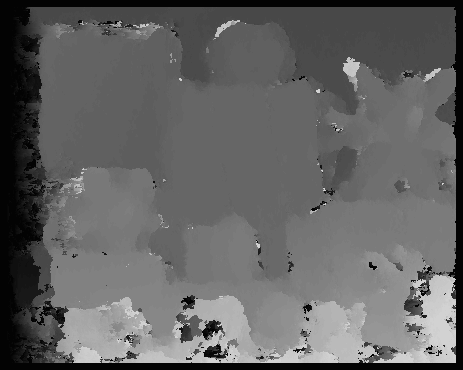}}\hfill
	\vspace{-21pt}
	\subfigure[(a) image 1]
	{\includegraphics[width=0.140\linewidth]{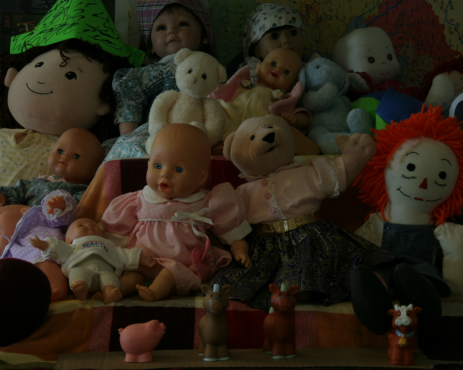}}\hfill
	\subfigure[(b) image 2]
	{\includegraphics[width=0.140\linewidth]{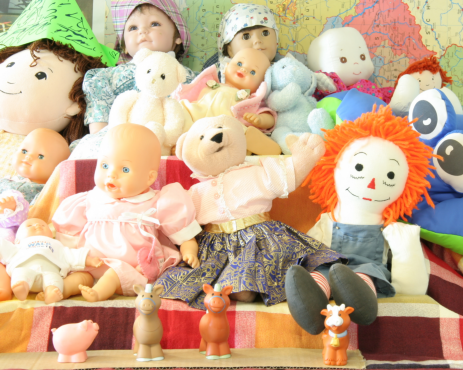}}\hfill
	\subfigure[(c) ANCC]
	{\includegraphics[width=0.140\linewidth]{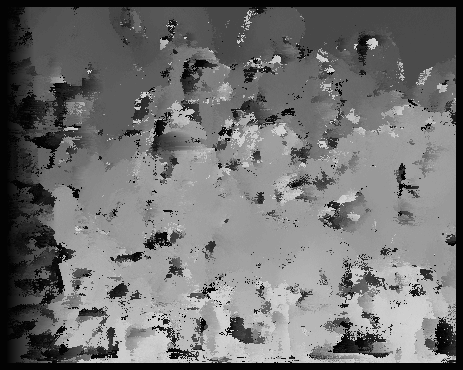}}\hfill
	\subfigure[(d) SIFT]
	{\includegraphics[width=0.140\linewidth]{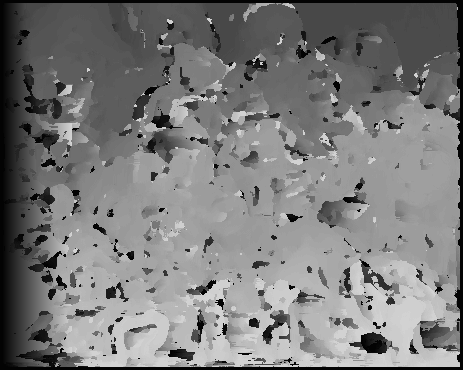}}\hfill
	\subfigure[(e) LSS]
	{\includegraphics[width=0.140\linewidth]{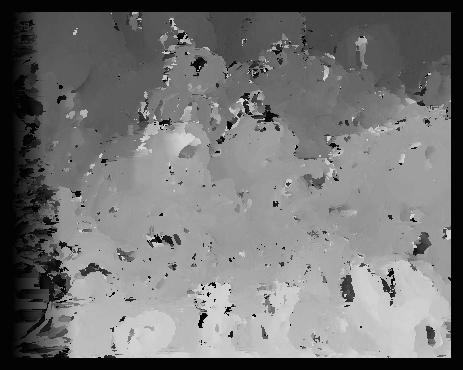}}\hfill
	\subfigure[(f) DASC]
	{\includegraphics[width=0.140\linewidth]{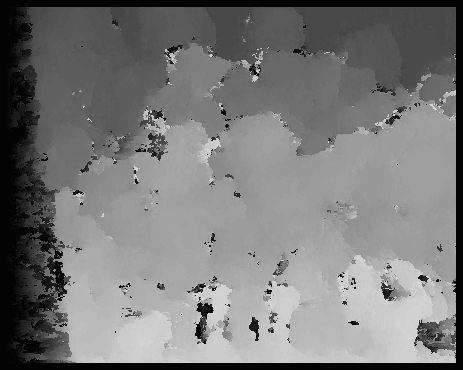}}\hfill
	\subfigure[(g) DeSCA]
	{\includegraphics[width=0.140\linewidth]{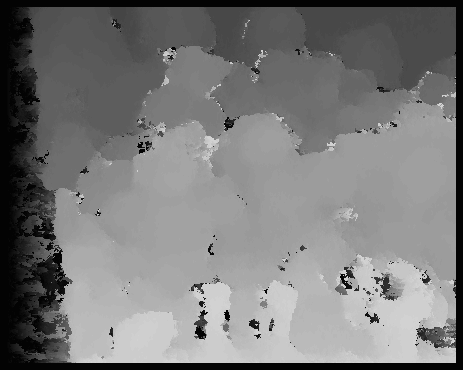}}\hfill
	\vspace{-12pt}
	\caption{Comparison of disparity estimations for \emph{Moebius} and \emph{Dolls} image pairs
		across illumination combination `1/3' and exposure combination `0/2', respectively. Compared to other methods,
		DeSCA estimates more accurate and edge-preserved disparity maps.}\label{img:8}\vspace{-10pt}
\end{figure}
\begin{figure}[!t]
	\centering
	\renewcommand{\thesubfigure}{}
	\subfigure[(a)]
	{\includegraphics[width=0.25\linewidth]{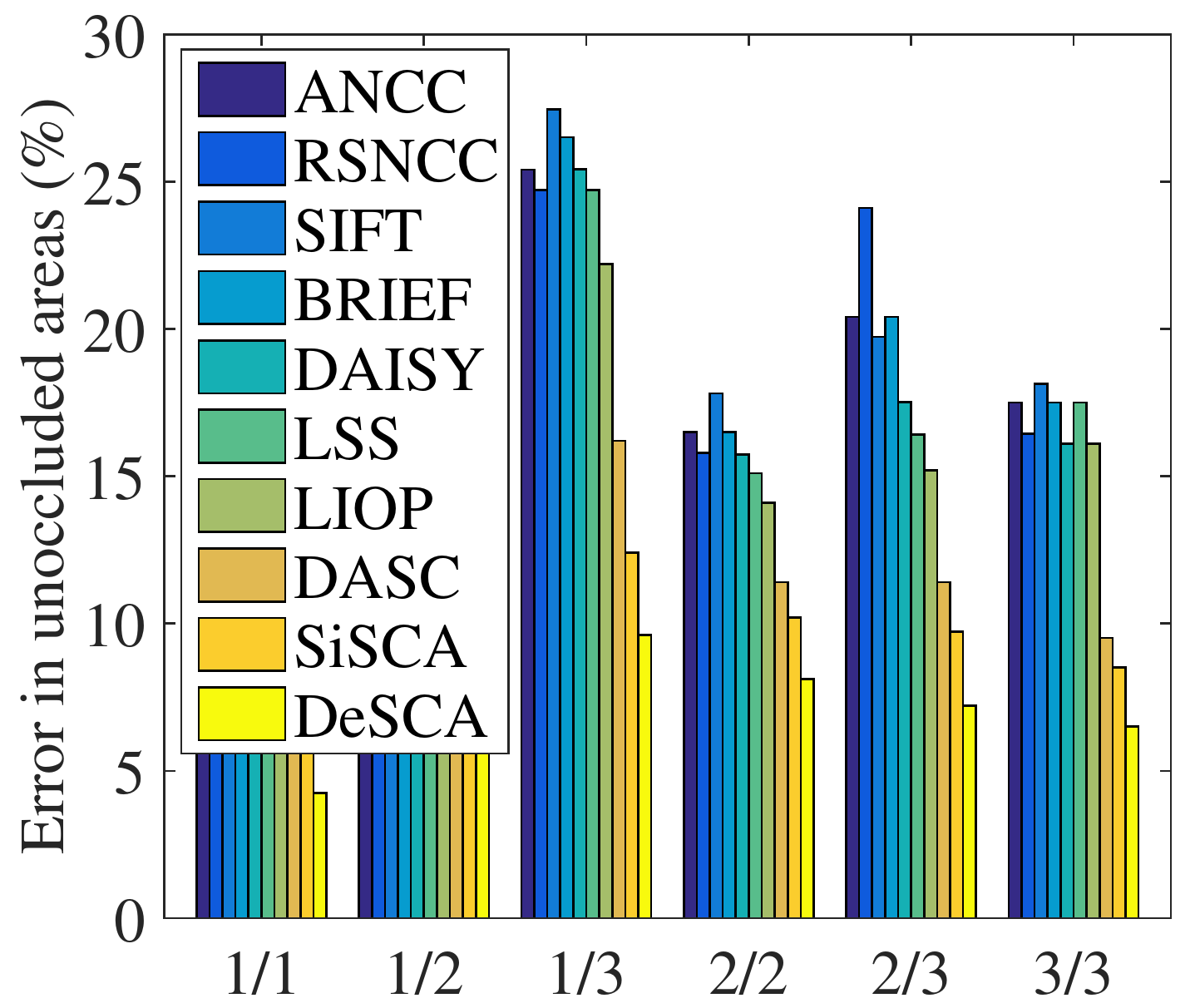}}\hfill
	\subfigure[(b)]
	{\includegraphics[width=0.25\linewidth]{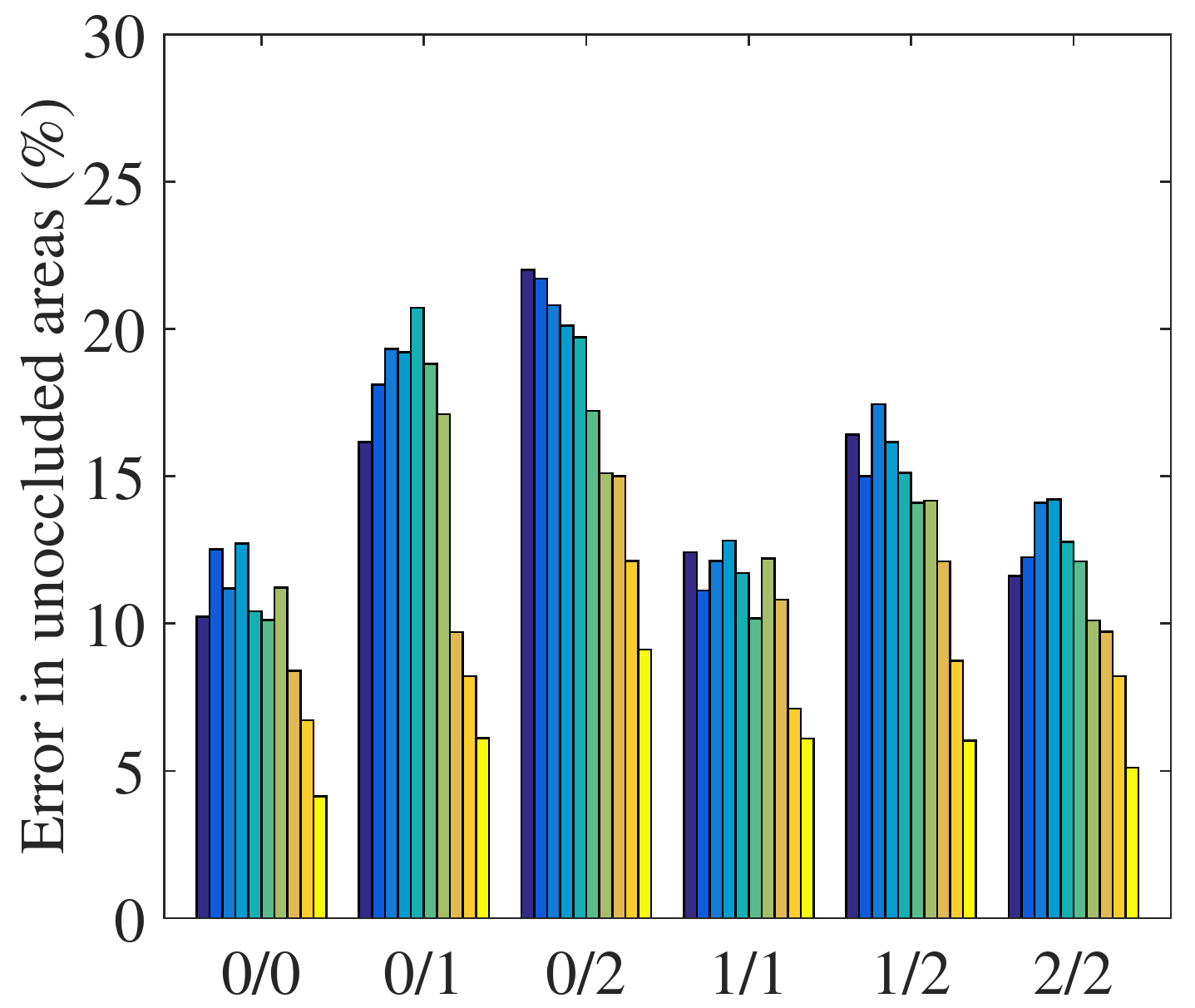}}\hfill
	\subfigure[(c)]
	{\includegraphics[width=0.25\linewidth]{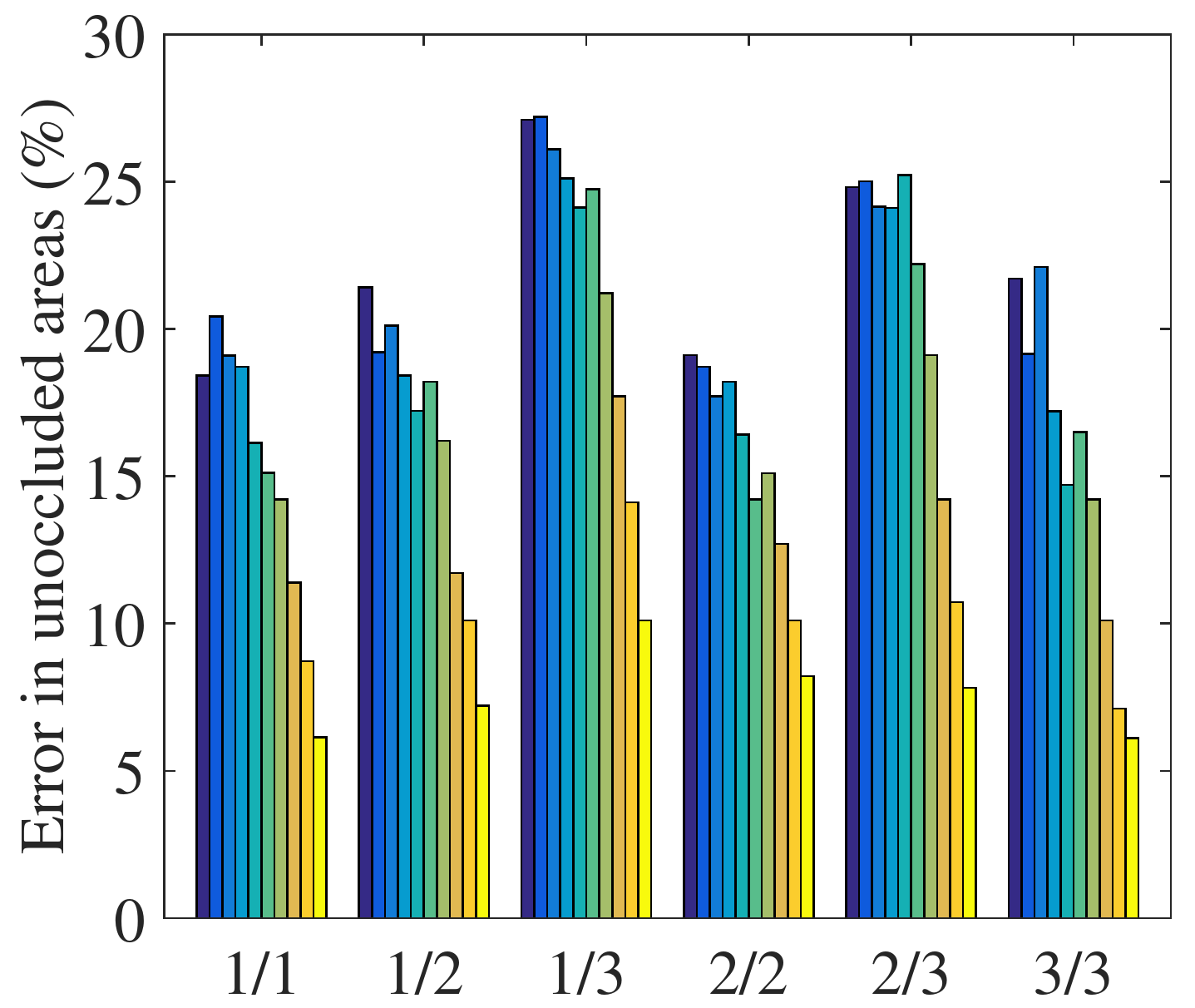}}\hfill
	\subfigure[(d)]
	{\includegraphics[width=0.25\linewidth]{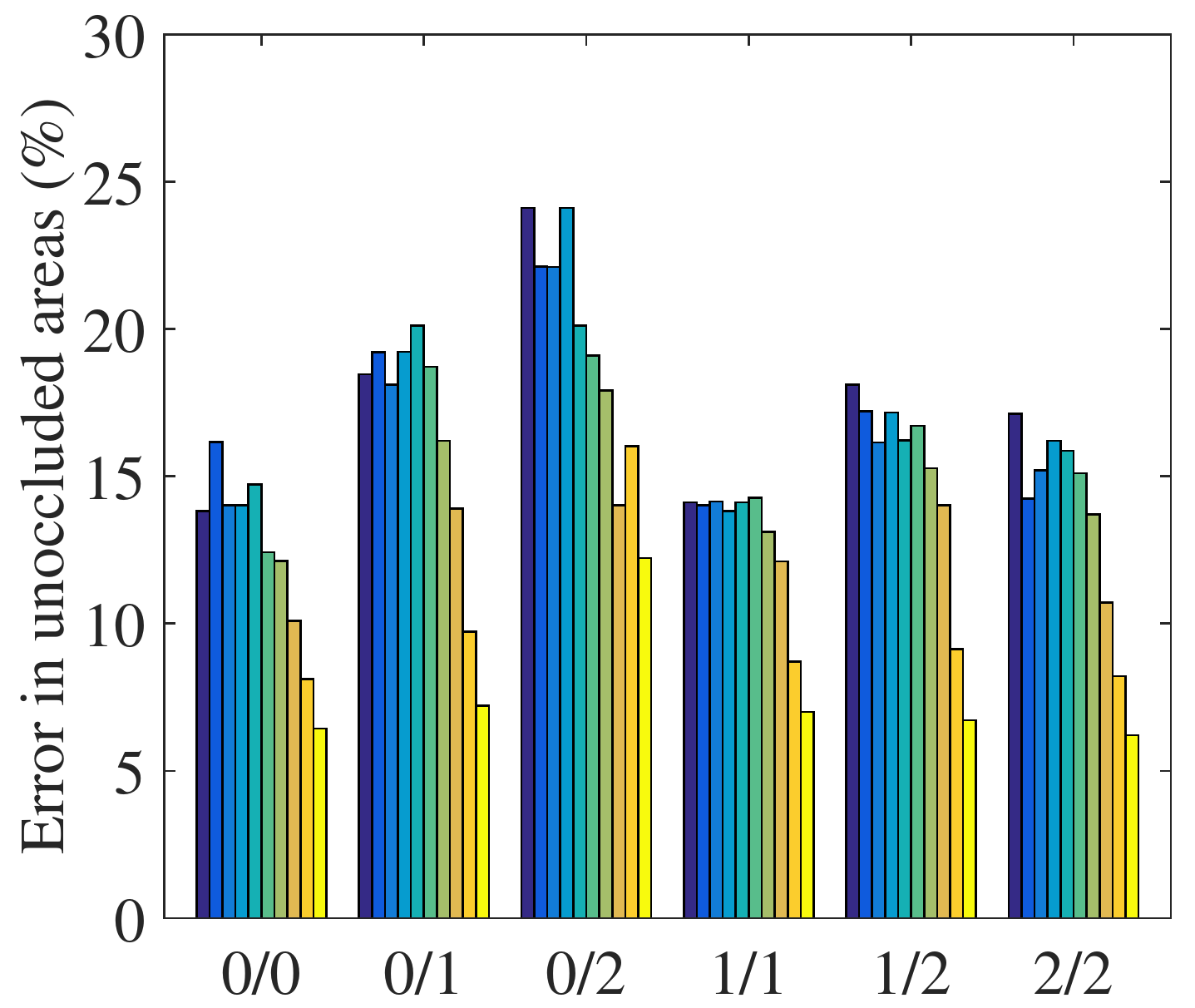}}\hfill
	\vspace{-12pt}
	\caption{Average bad-pixel error rate on the Middlebury benchmark \cite{middlebury} with illumination and exposure variations.
		Optimization was done by GC in (a), (b), and by WTA in (c), (d).
		DeSCA descriptor shows the best performance with the lowest error rate.}\label{img:9}\vspace{-10pt}
\end{figure}

\subsection{Parameter Evaluation}\label{sec:52}
The matching performance of DeSCA is exhibited in
\figref{img:7} for varying parameter values, including support window size $M_\mathcal{R}$,
number of log-polar circular points $N_\rho \times N_\theta$, number of random samples $N_K$,
and levels of the circular spatial pyramid $N_S$. Note that $N_O = N_S$. 
Especially, \figref{img:7}(c), (d) prove the effectiveness of self-convolutional activations and deep architectures of DeSCA.
For a quantitative analysis, we measured the average bad-pixel
error rate on the Middlebury benchmark \cite{middlebury}.
With a larger support window $M_\mathcal{R}$,
the matching quality improves rapidly until about $9 \times 9$.
$N_\rho \times N_\theta$ influences the performance of circular pooling,
which is found to plateau at $4 \times 16$.
Using a larger number of random samples $N_K$ yields better performance since the descriptor
encodes more information. The level of circular spatial pyramid $N_S$ also
affects the amount of encoding in DeSCA.
Based on these experiments, we set $N_K=32$ and $N_S=3$ in consideration of efficiency and robustness. \vspace{-5pt}
\begin{figure}[t!]
	\centering
	\renewcommand{\thesubfigure}{}
	\subfigure[]
	{\includegraphics[width=0.140\linewidth]{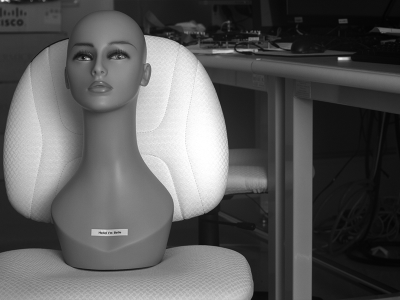}}\hfill
	\subfigure[]
	{\includegraphics[width=0.140\linewidth]{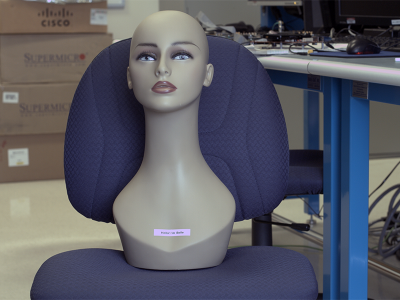}}\hfill
	\subfigure[]
	{\includegraphics[width=0.140\linewidth]{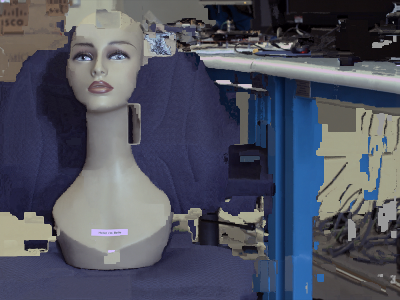}}\hfill
	\subfigure[]
	{\includegraphics[width=0.140\linewidth]{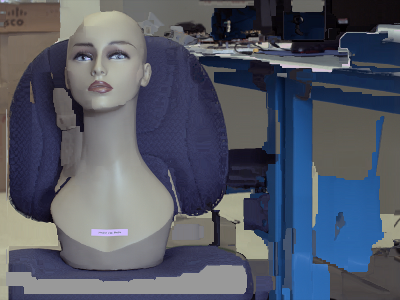}}\hfill
	\subfigure[]
	{\includegraphics[width=0.140\linewidth]{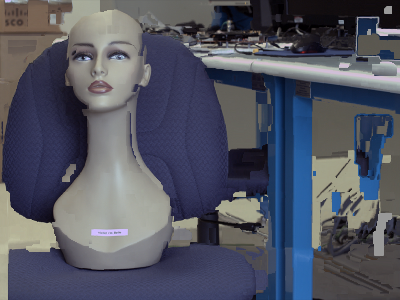}}\hfill
	\subfigure[]
	{\includegraphics[width=0.140\linewidth]{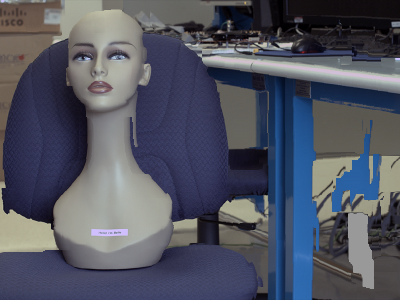}}\hfill
	\subfigure[]
	{\includegraphics[width=0.140\linewidth]{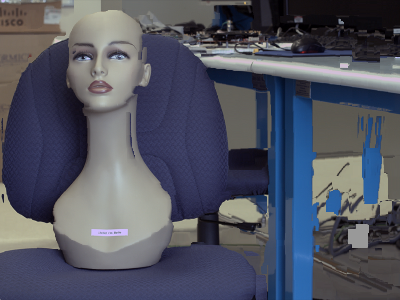}}\hfill
	\vspace{-21pt}
	\subfigure[]
	{\includegraphics[width=0.140\linewidth]{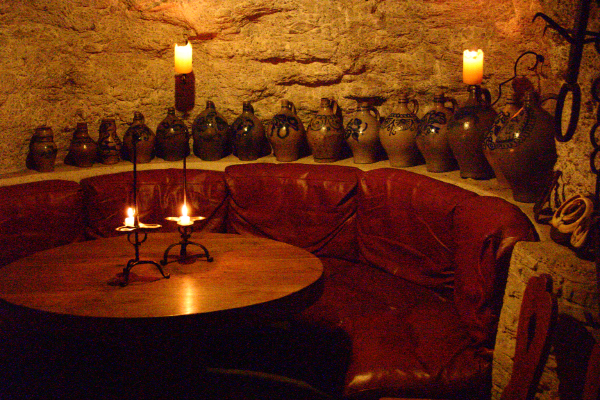}}\hfill
	\subfigure[]
	{\includegraphics[width=0.140\linewidth]{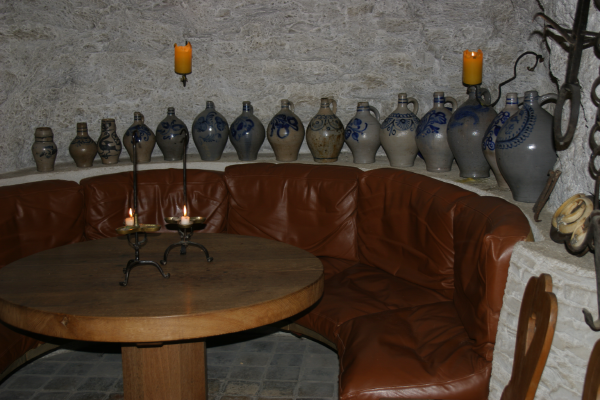}}\hfill
	\subfigure[]
	{\includegraphics[width=0.140\linewidth]{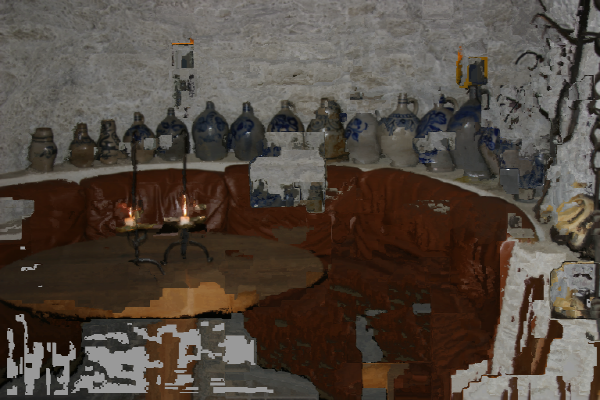}}\hfill
	\subfigure[]
	{\includegraphics[width=0.140\linewidth]{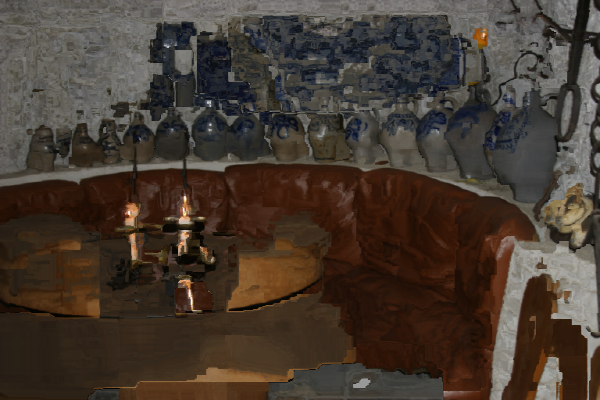}}\hfill
	\subfigure[]
	{\includegraphics[width=0.140\linewidth]{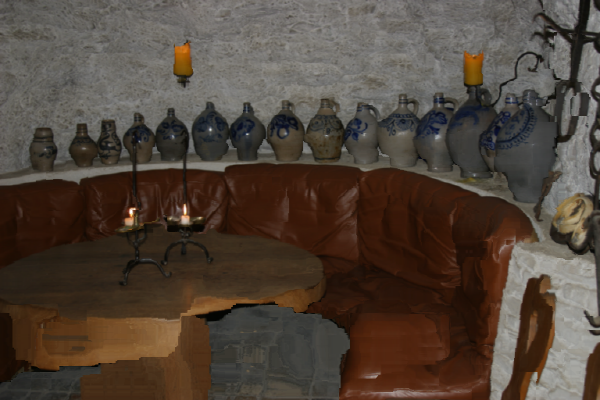}}\hfill
	\subfigure[]
	{\includegraphics[width=0.140\linewidth]{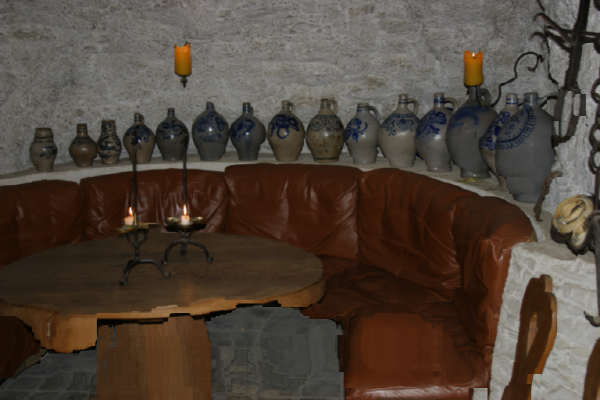}}\hfill
	\subfigure[]
	{\includegraphics[width=0.140\linewidth]{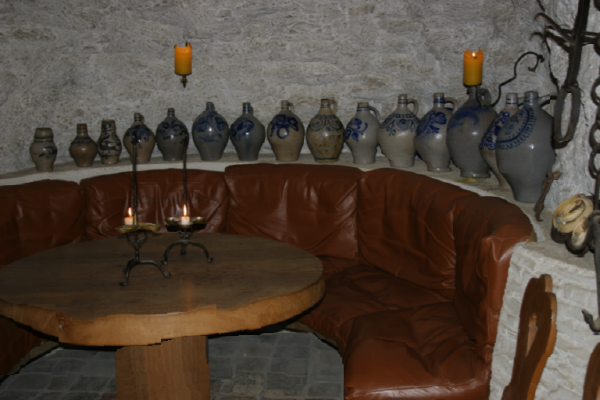}}\hfill
	\vspace{-21pt}
	\subfigure[]
	{\includegraphics[width=0.140\linewidth]{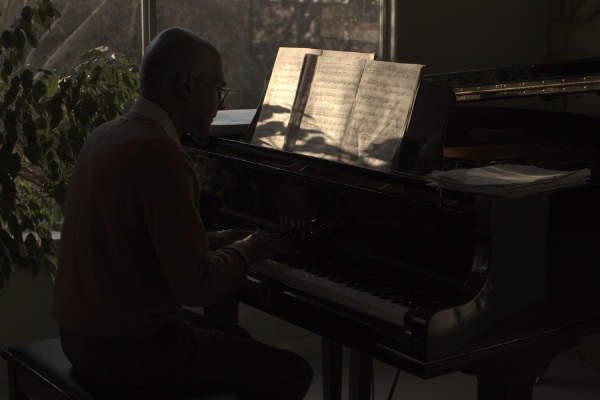}}\hfill
	\subfigure[]
	{\includegraphics[width=0.140\linewidth]{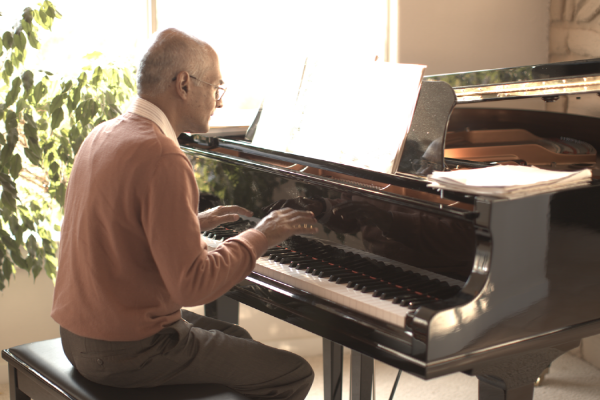}}\hfill
	\subfigure[]
	{\includegraphics[width=0.140\linewidth]{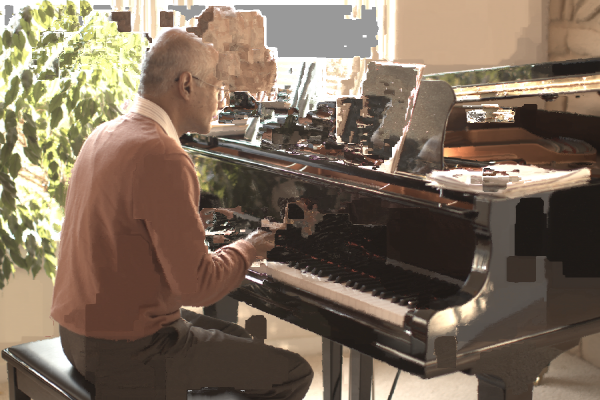}}\hfill
	\subfigure[]
	{\includegraphics[width=0.140\linewidth]{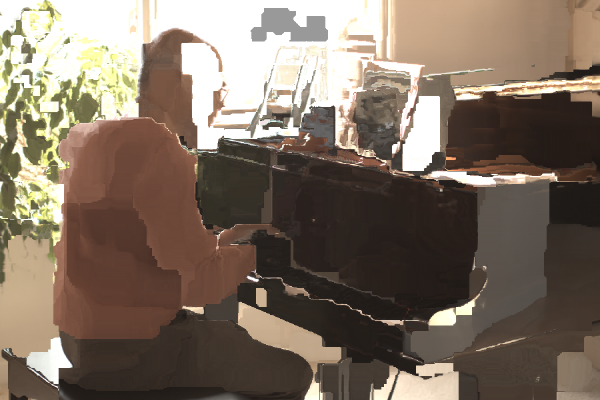}}\hfill
	\subfigure[]
	{\includegraphics[width=0.140\linewidth]{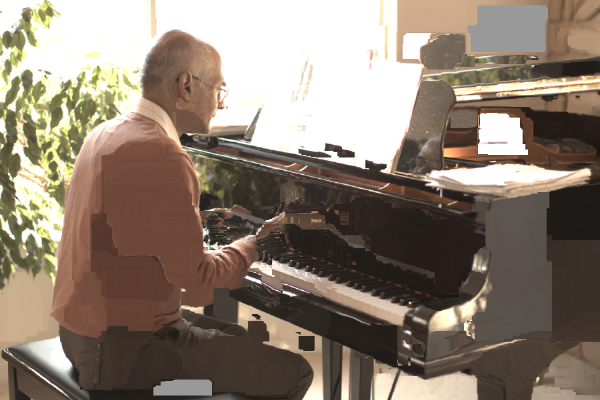}}\hfill
	\subfigure[]
	{\includegraphics[width=0.140\linewidth]{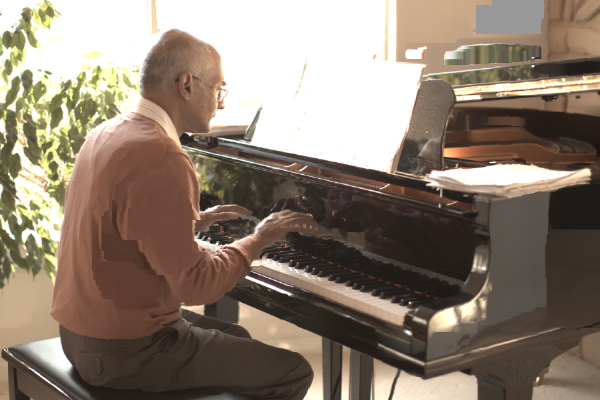}}\hfill
	\subfigure[]
	{\includegraphics[width=0.140\linewidth]{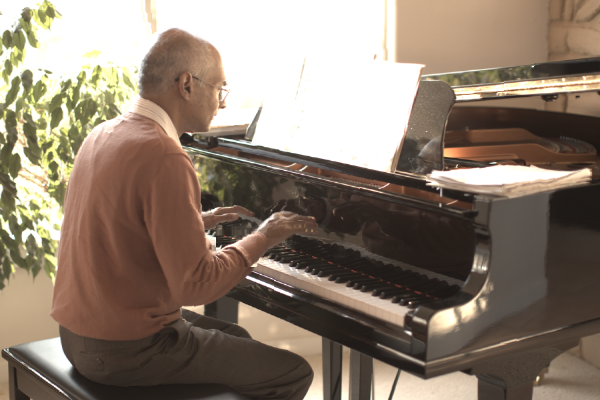}}\hfill
	\vspace{-21pt}
	\subfigure[(a) image 1]
	{\includegraphics[width=0.140\linewidth]{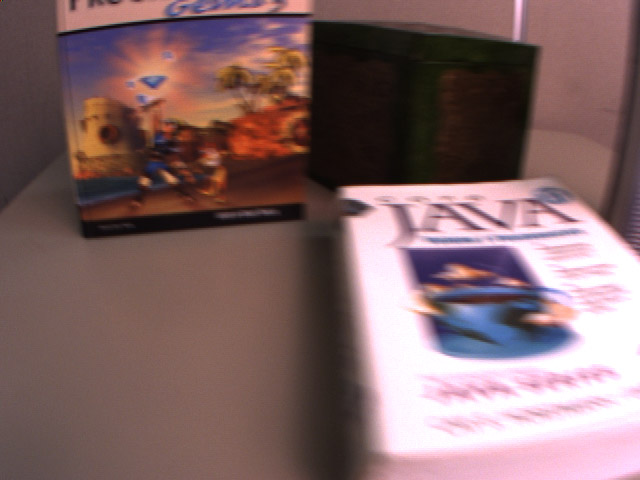}}\hfill
	\subfigure[(b) image 2]
	{\includegraphics[width=0.140\linewidth]{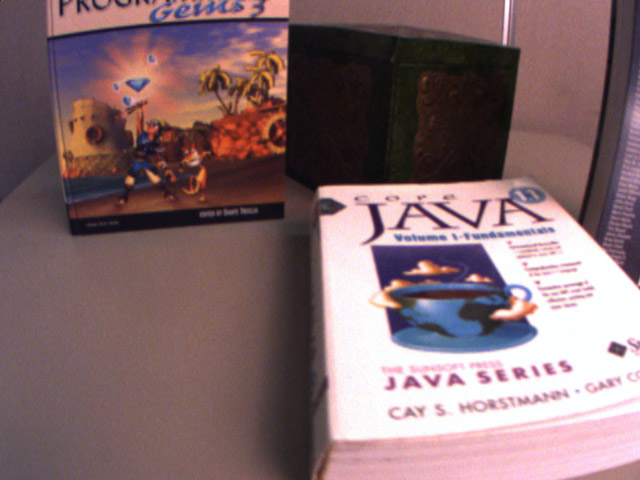}}\hfill
	\subfigure[(c) BRIEF]
	{\includegraphics[width=0.140\linewidth]{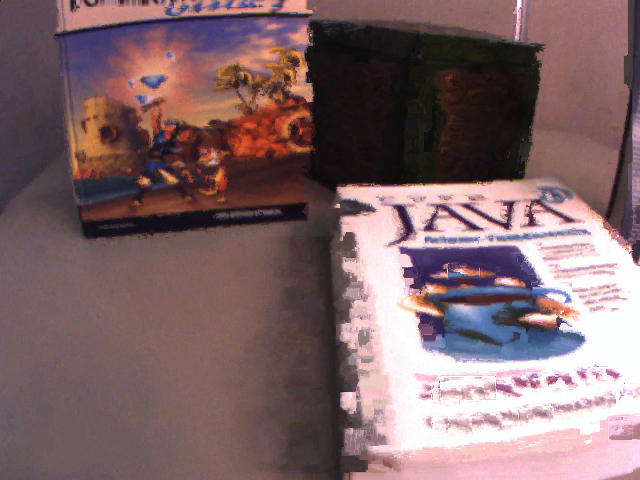}}\hfill
	\subfigure[(d) LSS]
	{\includegraphics[width=0.140\linewidth]{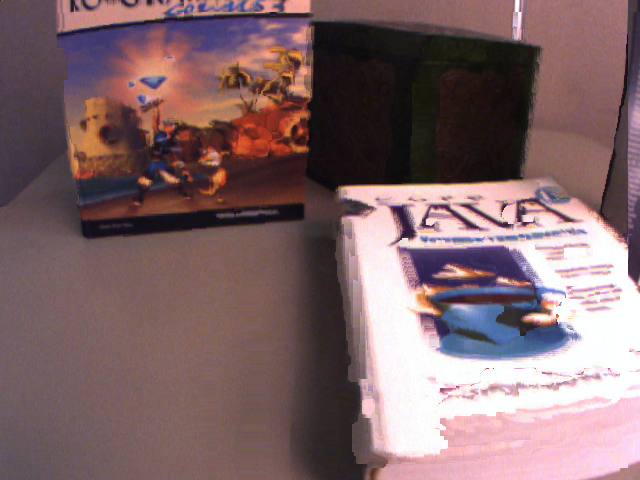}}\hfill
	\subfigure[(e) DASC]
	{\includegraphics[width=0.140\linewidth]{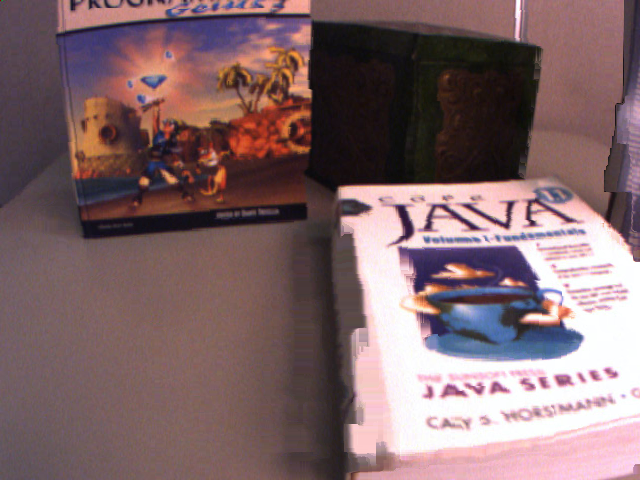}}\hfill
	\subfigure[(f) SiSCA]
	{\includegraphics[width=0.140\linewidth]{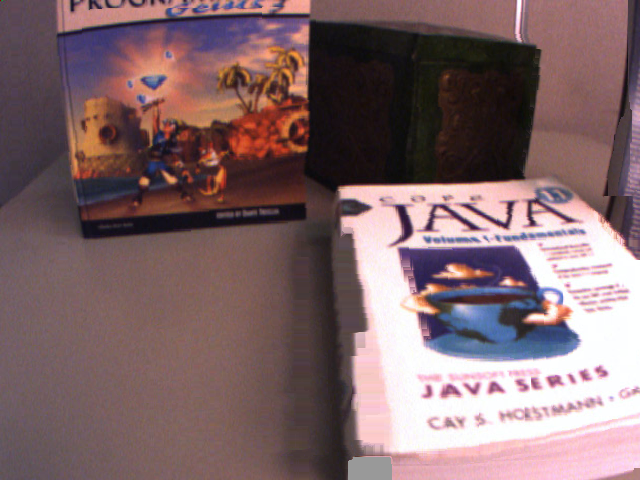}}\hfill
	\subfigure[(g) DeSCA]
	{\includegraphics[width=0.140\linewidth]{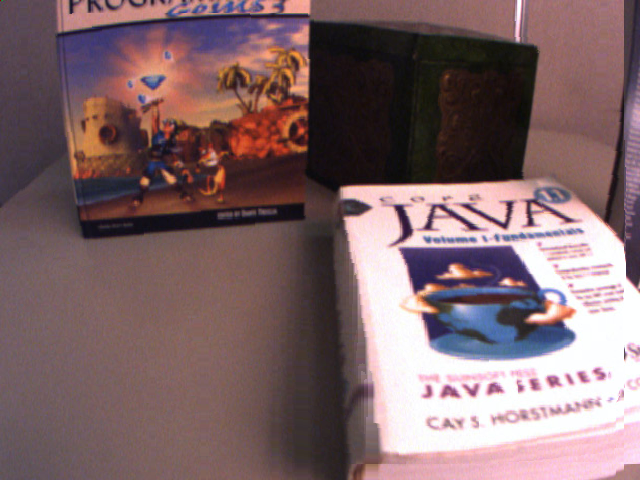}}\hfill
	\vspace{-12pt}
	\caption{Dense correspondence evaluations for
		(from top to bottom) RGB-NIR, flash-noflash, different exposures, and blurred-sharp images.
		Compared to others, DeSCA estimates more reliable
		dense correspondences for challenging cross-modal pairs.}\label{img:10}\vspace{-5pt}
\end{figure}
\begin{table}[t]
	\centering
	\begin{tabular}{ >{\raggedright}m{0.16\linewidth}
			>{\centering}m{0.09\linewidth}  >{\centering}m{0.09\linewidth}
			>{\centering}m{0.09\linewidth}  >{\centering}m{0.09\linewidth}
			>{\centering}m{0.09\linewidth}  >{\centering}m{0.09\linewidth}
			>{\centering}m{0.09\linewidth}  >{\centering}m{0.09\linewidth}  }
		\hlinewd{0.8pt}
		\multirow{2}{*}{$\;$Methods} & \multicolumn{4}{ c }{WTA optimization} & \multicolumn{4}{ c }{SF optimization \cite{Liu11}} \tabularnewline
		\cline{2-9}
		&RGB-NIR &{\footnotesize flash-noflash} &diff. expo. &blur-sharp
		&RGB-NIR &{\footnotesize flash-noflash} &diff. expo. &blur-sharp \tabularnewline
		\hline
		\hline
		$\;$ANCC \cite{Heo11} &23.21 &20.42 &25.19 &26.14 &18.45 &14.14 &11.96 &19.24 \tabularnewline
		$\;$RSNCC \cite{Shen14} &27.51 &25.12 &18.21 &27.91 &13.41 &15.87 &9.15 &18.21 \tabularnewline
		$\;$SIFT \cite{Lowe04} &24.11 &18.72 &19.42 &27.18 &18.51 &11.06 &14.87 &20.78 \tabularnewline
		$\;$DAISY \cite{Tola10} &27.61 &26.30 &20.72 &27.41 &20.42 &10.84 &12.71 &22.91 \tabularnewline
		$\;$BRIEF \cite{Calonder11} &29.14 &18.29 &17.13 &26.43  &17.54 &9.21 &9.54 &19.72 \tabularnewline
		$\;$LSS \cite{Schechtman07} &27.82 &19.18 &18.21 &26.14 &16.14 &11.88 &9.11 &18.51 \tabularnewline
		$\;$LIOP \cite{Wang11} &24.42 &16.42 &14.22 &20.42 &15.32 &11.42 &10.22 &17.12 \tabularnewline
		$\;$DASC \cite{Kim15} &14.51 &13.24 &10.32 &16.42 &13.42 &7.11 &7.21 &11.21 \tabularnewline
		\hline
		$\;$\textbf{SiSCA}\cellcolor{blue!5} &\textbf{10.12}\cellcolor{blue!5} &\textbf{10.12}\cellcolor{blue!5} &\textbf{8.22}\cellcolor{blue!5} &\textbf{14.22}\cellcolor{blue!5} &\textbf{9.12}\cellcolor{blue!5} &\textbf{6.18}\cellcolor{blue!5} &\textbf{5.22}\cellcolor{blue!5} &\textbf{9.12}\cellcolor{blue!5} \tabularnewline
		$\;$\textbf{DeSCA}\cellcolor{blue!5} &\textbf{8.12}\cellcolor{blue!5} &\textbf{8.22}\cellcolor{blue!5} &\textbf{6.72}\cellcolor{blue!5} &\textbf{13.28}\cellcolor{blue!5} &\textbf{7.62}\cellcolor{blue!5} &\textbf{5.12}\cellcolor{blue!5} &\textbf{4.72}\cellcolor{blue!5} &\textbf{8.01}\cellcolor{blue!5} \tabularnewline
		\hlinewd{0.8pt}
	\end{tabular}\vspace{+3pt}
	\caption{Comparison of quantitative evaluation on cross-modal benchmark.}\label{tab:1}\vspace{-20pt}
\end{table}

\subsection{Middlebury Stereo Benchmark}\label{sec:53}
We evaluated DeSCA on the Middlebury stereo
benchmark \cite{middlebury}, which contains illumination and exposure variations.
In the experiments, the illumination (exposure) combination `1/3' indicates that two images were captured
under the $1^{st}$ and $3^{rd}$ illumination (exposure) conditions. For a quantitative evaluation,
we measured the bad-pixel error rate in non-occluded areas of disparity maps \cite{middlebury}. 

\figref{img:8} shows the disparity maps estimated
under severe illumination and exposure variations
%by varying the cost functions
with winner-takes-all (WTA) optimization.
\figref{img:9} displays the average bad-pixel
error rates of disparity maps obtained under illumination or exposure variations,
with graph-cut (GC) \cite{Boykov01} and WTA optimization.
Area-based approaches (ANCC \cite{Heo11} and RSNCC \cite{Shen14})
are sensitive to severe radiometric variations,
especially when local variations occur frequently.
Feature descriptor-based methods (SIFT \cite{Lowe04}, DAISY \cite{Tola10}, BRIEF
\cite{Calonder11}, LSS \cite{Schechtman07}, and DASC \cite{Kim15}) perform better than the area-based
approaches, but they also provide limited performance.
Our DeSCA descriptor achieves the best results both quantitatively and qualitatively.
Compared to SiSCA descriptor, the performance of DeSCA descriptor is highly improved, 
where the performance benefits of the deep architecture are apparent.\vspace{-5pt}
\begin{figure}[t]
	\centering
	\renewcommand{\thesubfigure}{}
	\subfigure[]
	{\includegraphics[width=0.140\linewidth]{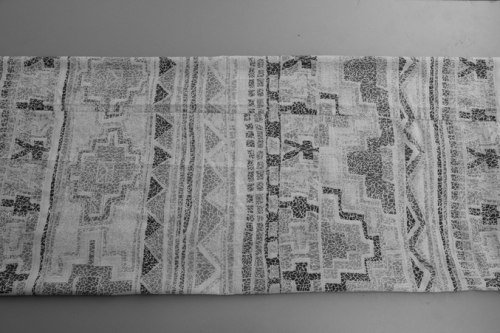}}\hfill
	\subfigure[]
	{\includegraphics[width=0.140\linewidth]{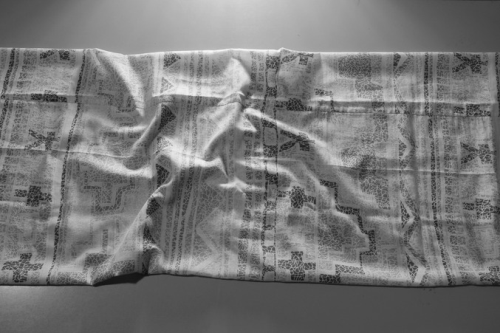}}\hfill
	\subfigure[]
	{\includegraphics[width=0.140\linewidth]{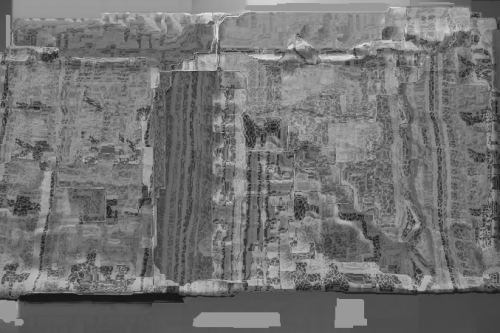}}\hfill
	\subfigure[]
	{\includegraphics[width=0.140\linewidth]{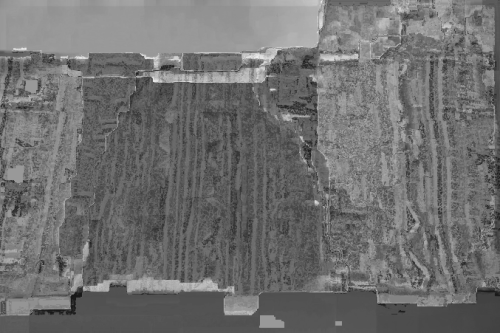}}\hfill
	\subfigure[]
	{\includegraphics[width=0.140\linewidth]{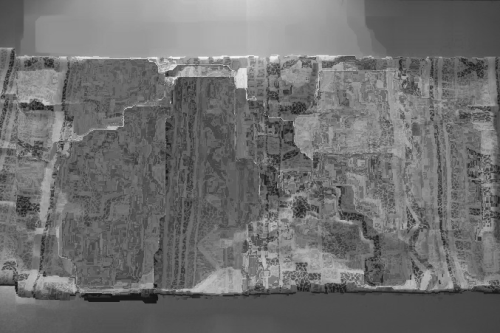}}\hfill
	\subfigure[]
	{\includegraphics[width=0.140\linewidth]{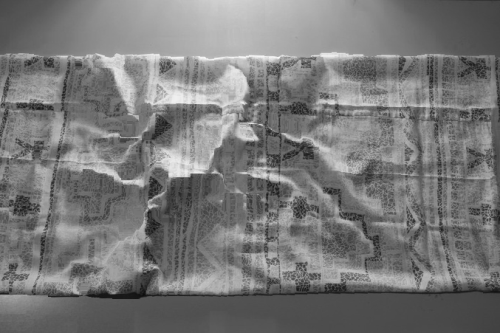}}\hfill
	\subfigure[]
	{\includegraphics[width=0.140\linewidth]{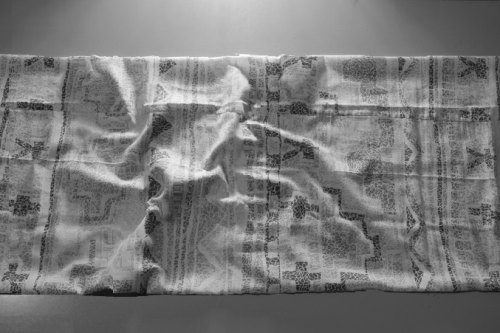}}\hfill
	\vspace{-21pt}
	\subfigure[]
	{\includegraphics[width=0.140\linewidth]{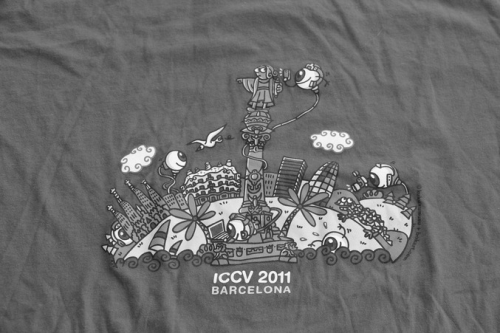}}\hfill
	\subfigure[]
	{\includegraphics[width=0.140\linewidth]{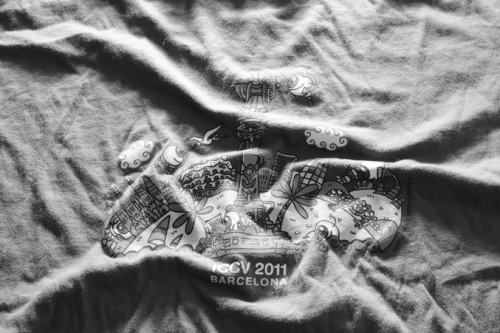}}\hfill
	\subfigure[]
	{\includegraphics[width=0.140\linewidth]{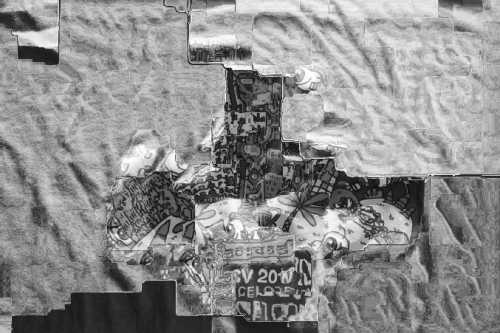}}\hfill
	\subfigure[]
	{\includegraphics[width=0.140\linewidth]{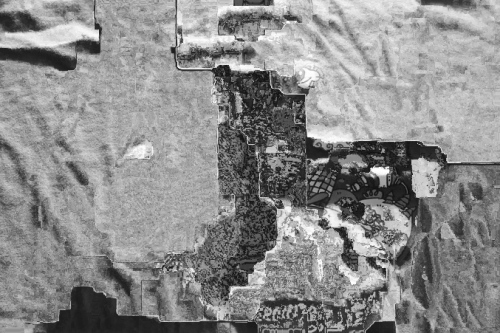}}\hfill
	\subfigure[]
	{\includegraphics[width=0.140\linewidth]{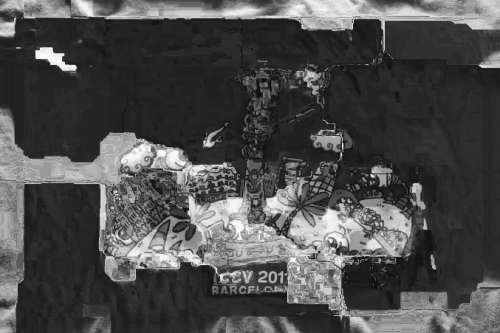}}\hfill
	\subfigure[]
	{\includegraphics[width=0.140\linewidth]{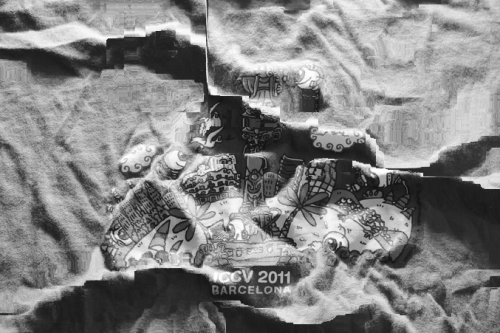}}\hfill
	\subfigure[]
	{\includegraphics[width=0.140\linewidth]{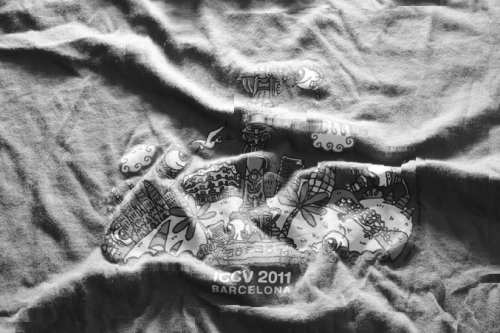}}\hfill
	\vspace{-21pt}
	\subfigure[]
	{\includegraphics[width=0.140\linewidth]{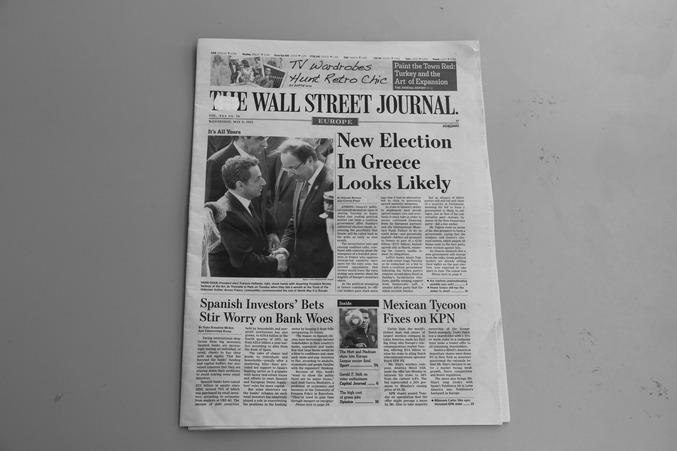}}\hfill
	\subfigure[]
	{\includegraphics[width=0.140\linewidth]{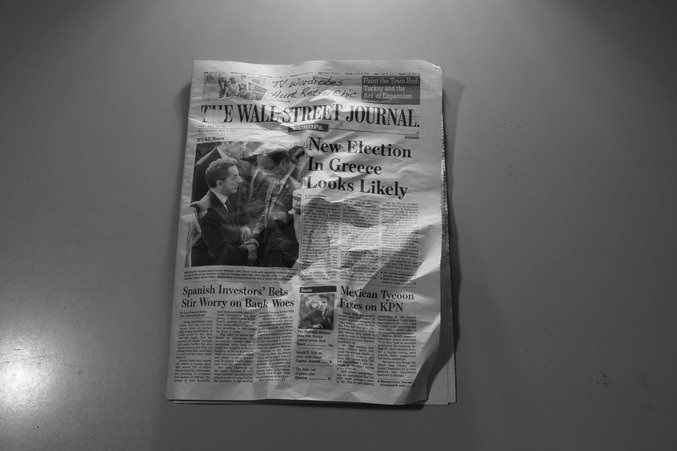}}\hfill
	\subfigure[]
	{\includegraphics[width=0.140\linewidth]{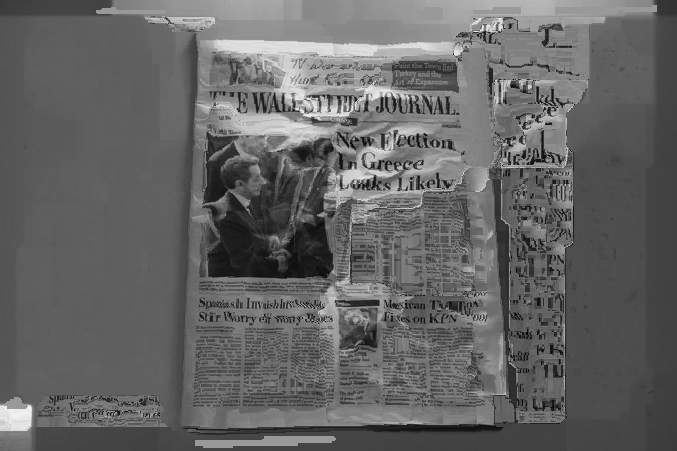}}\hfill
	\subfigure[]
	{\includegraphics[width=0.140\linewidth]{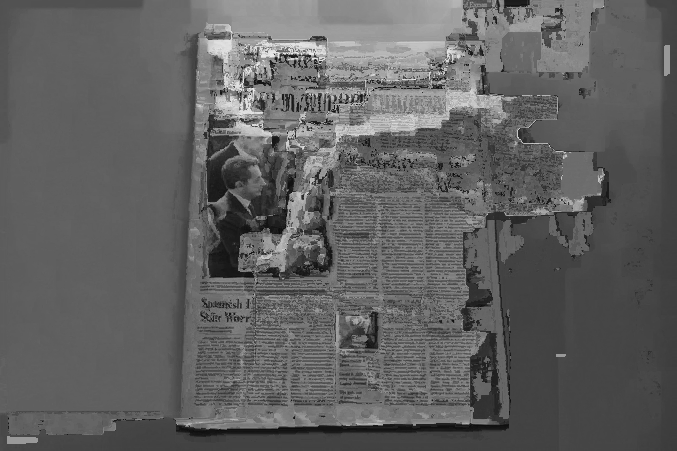}}\hfill
	\subfigure[]
	{\includegraphics[width=0.140\linewidth]{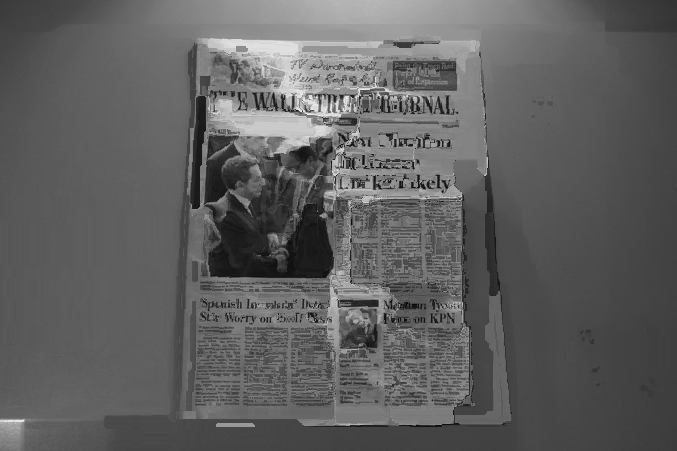}}\hfill
	\subfigure[]
	{\includegraphics[width=0.140\linewidth]{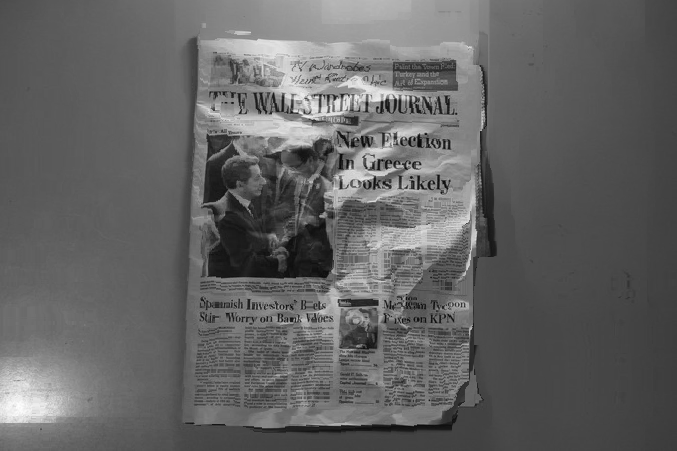}}\hfill
	\subfigure[]
	{\includegraphics[width=0.140\linewidth]{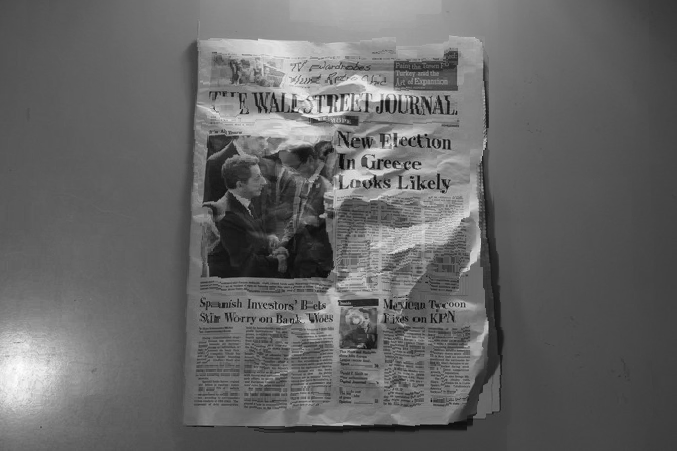}}\hfill
	\vspace{-21pt}
	\subfigure[(a) image 1]
	{\includegraphics[width=0.140\linewidth]{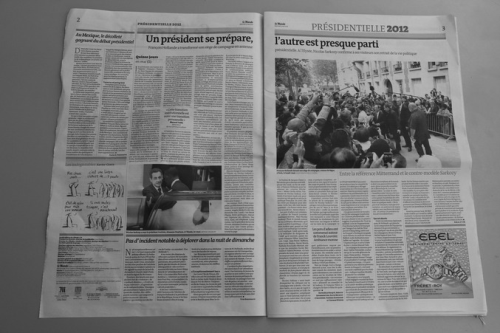}}\hfill
	\subfigure[(b) image 2]
	{\includegraphics[width=0.140\linewidth]{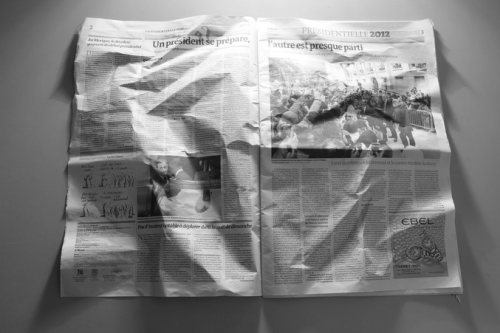}}\hfill
	\subfigure[(c) DAISY]
	{\includegraphics[width=0.140\linewidth]{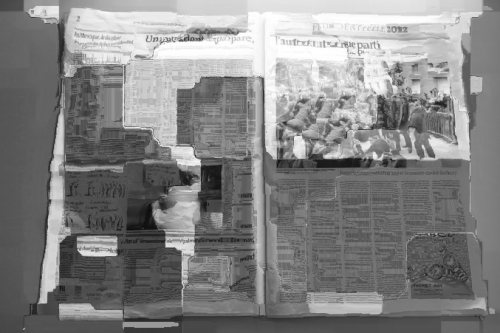}}\hfill
	\subfigure[(d) BRIEF]
	{\includegraphics[width=0.140\linewidth]{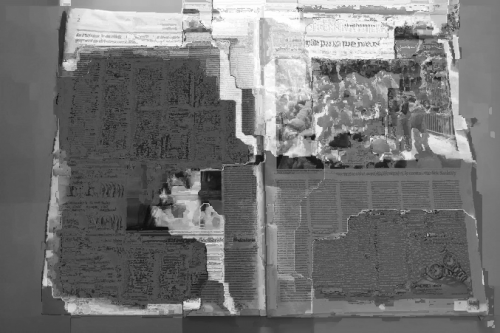}}\hfill
	\subfigure[(e) LSS]
	{\includegraphics[width=0.140\linewidth]{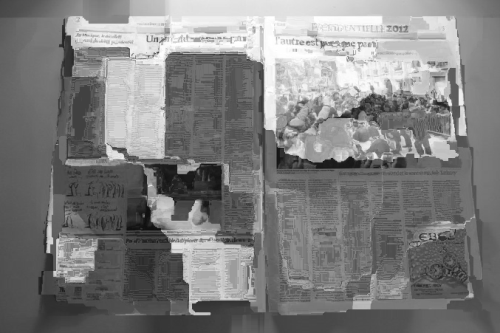}}\hfill
	\subfigure[(f) DaLI]
	{\includegraphics[width=0.140\linewidth]{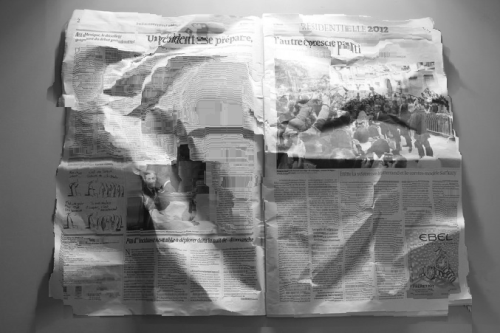}}\hfill
	\subfigure[(g) DeSCA]
	{\includegraphics[width=0.140\linewidth]{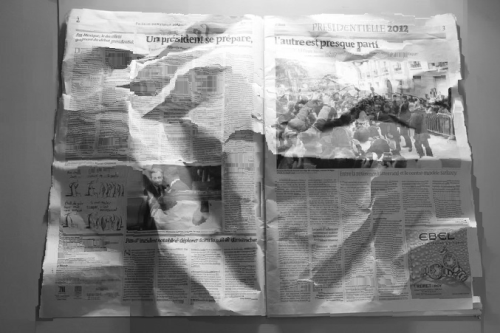}}\hfill
	\vspace{-12pt}
	\caption{Dense correspondence comparisons for images with different illumination conditions and non-rigid image deformations \cite{Simo-Serra15}.
		Compared to other approaches, DeSCA provides more accurate dense correspondence estimates with reduced artifacts.}\label{img:11}\vspace{-5pt}
\end{figure}
\begin{table}[t]
	\centering
	\begin{tabular}{ >{\raggedright}m{0.16\linewidth}
			>{\centering}m{0.09\linewidth}  >{\centering}m{0.09\linewidth}
			>{\centering}m{0.09\linewidth}  >{\centering}m{0.09\linewidth} }
		\hlinewd{0.8pt}
		$\;$Methods &def. &illum. &def./ illum. &aver. \tabularnewline
		\hline
		\hline
		$\;$SIFT \cite{Lowe04} &45.15 &40.81 &47.51 &44.49 \tabularnewline
		$\;$DAISY \cite{Tola10} &43.98 &42.72 &43.42 &43.37 \tabularnewline
		$\;$BRIEF \cite{Calonder11} &41.51 &37.14 &41.35 &40 \tabularnewline
		$\;$LSS \cite{Schechtman07} &40.81 &39.54 &40.11 &40.12 \tabularnewline
		$\;$LIOP \cite{Wang11} &28.72 &31.72 &30.21 &30.22 \tabularnewline
		$\;$DaLI \cite{Simo-Serra15} &27.12 &27.31 &27.99 &27.47 \tabularnewline
		$\;$DASC \cite{Kim15} &26.21 &24.83 &27.51 &26.18 \tabularnewline
		\hline
		$\;$\textbf{SiSCA}\cellcolor{blue!5} &\textbf{23.42}\cellcolor{blue!5} &\textbf{22.21}\cellcolor{blue!5} &\textbf{24.17}\cellcolor{blue!5} &\textbf{23.27}\cellcolor{blue!5} \tabularnewline
		$\;$\textbf{DeSCA}\cellcolor{blue!5} &\textbf{20.14}\cellcolor{blue!5} &\textbf{20.72}\cellcolor{blue!5} &\textbf{21.87}\cellcolor{blue!5} &\textbf{20.91}\cellcolor{blue!5} \tabularnewline
		\hlinewd{0.8pt}
	\end{tabular}\vspace{+3pt}
	\caption{Average error rates on the DaLI benchmark.}\label{tab:2}\vspace{-20pt}
\end{table}

\subsection{Cross-modal and Cross-spectral Benchmark}\label{sec:54}
We evaluated DeSCA on a cross-modal and cross-spectral benchmark \cite{Kim15}
containing various kinds of image pairs, namely RGB-NIR, different exposures, flash-noflash, and blurred-sharp. Optimization for all descriptors and similarity measures
was done using WTA and SIFT flow (SF)
with hierarchical dual-layer belief propagation
\cite{Liu11}, for which the code is publicly available.
Sparse ground truths for those images are used for error measurement as done in \cite{Kim15}.

\figref{img:10} provides a qualitative comparison of the DeSCA descriptor
to other state-of-the-art approaches. As already described in the literature \cite{Shen14},
gradient-based approaches such as SIFT \cite{Lowe04} and DAISY
\cite{Tola10} have shown limited performance for RGB-NIR pairs where
gradient reversals and inversions frequently appear. BRIEF
\cite{Calonder11} cannot deal with noisy regions and modality-based
appearance differences since it is formulated on pixel differences only.
Unlike these approaches, LSS \cite{Schechtman07} and DASC \cite{Kim15}
consider local self-similarities, but LSS is lacking in
discriminative power for dense matching. DASC also exhibits limited performance.
Compared to those methods, the DeSCA displays better correspondence estimation.
We also performed a quantitative evaluation with results listed in \tabref{tab:1}, which also
clearly demonstrates the effectiveness of DeSCA.\vspace{-5pt}

\subsection{DaLI Benchmark}\label{sec:55}
We also evaluated DeSCA on a recent,
publicly available dataset featuring challenging non-rigid deformations
and very severe illumination changes \cite{Simo-Serra15}.
\figref{img:11} presents dense correspondence estimates for this benchmark \cite{Simo-Serra15}.
A quantitative evaluation is given in \tabref{tab:2} using ground truth feature points sparsely extracted for each image,
although DeSCA is designed to estimate dense correspondences.
As expected, conventional gradient-based and intensity comparison-based feature descriptors,
including SIFT \cite{Lowe04}, DAISY \cite{Tola10}, and BRIEF \cite{Calonder11},
do not provide reliable correspondence performance.
LSS \cite{Schechtman07} and DASC \cite{Kim15} exhibit relatively
high performance for illumination changes, but are limited on non-rigid deformations.
LIOP \cite{Wang11} provides robustness to radiometric variations, but is sensitive to non-rigid deformations.
Although DaLI \cite{Simo-Serra15} provides robust correspondences, it requires considerable computation for dense matching.
DeSCA offers greater discriminative power
as well as more robustness to non-rigid deformations in comparison to the state-of-the-art cross-modality descriptors.\vspace{-5pt}
\begin{table}[t]
	\begin{center}
		\begin{tabular}{c>{\centering}m{0.11\linewidth}>{\centering}m{0.11\linewidth}>{\centering}m{0.11\linewidth}>{\centering}m{0.11\linewidth}>{\centering}m{0.11\linewidth}>{\centering}m{0.11\linewidth}>{\centering}m{0.11\linewidth}}
			\hlinewd{0.8pt}
			image size &SIFT &DAISY &LSS &DaLI &DASC &DeSCA* &DeSCA\dag \tabularnewline
			\hline
			\hline
			$463 \times 370$ &$130.3s$ &$2.5s$ &$31s$ &$352.2s$ &$2.7s$ &$193.2s$
			&$\mathbf{9.2s}$\cellcolor{blue!5} \tabularnewline
			\hlinewd{0.8pt}
		\end{tabular}
	\end{center}\vspace{-5pt}
	\caption{Computation speed of DeSCA and other state-of-the-art local and global descriptors. The brute-force and efficient implementations of DeSCA are denoted by * and \dag, respectively.}\label{tab:3}\vspace{-20pt}
\end{table}

\subsection{Computational Speed}\label{sec:56}
In \tabref{tab:3}, we compared the computational speed of DeSCA
to state-of-the-art local descriptor, namely DaLI \cite{Simo-Serra15}, and dense descriptors, namely DAISY \cite{Tola10}, LSS \cite{Schechtman07},
and DASC \cite{Kim15}.
Even though DeSCA needs more computational time compared to some previous dense descriptors,
it provides significantly improved matching performance as described previously. \vspace{-5pt}

\section{Conclusion}\label{sec:6}
The deep self-convolutional activations (DeSCA) descriptor was proposed
for establishing dense correspondences between images taken under different imaging modalities.
Its high performance in comparison to state-of-the-art cross-modality descriptors can be attributed
to its greater robustness to non-rigid deformations because of its effective pooling scheme, and more
importantly its heightened discriminative power from a more comprehensive representation of
self-similar structure and its formulation in a deep architecture. DeSCA was validated on an
extensive set of experiments that cover a broad range of cross-modal differences. 
In future work, thanks to the robustness to non-rigid deformations and high discriminative power, DeSCA can potentially benefit object detection and semantic segmentation. \vspace{-5pt}

\bibliographystyle{splncs}
\bibliography{egbib}
\end{document}